\documentclass[a4paper,12pt]{article}
\usepackage{amsmath}
\usepackage{amsfonts}
\usepackage{authblk}
\usepackage{array}
\usepackage{longtable}
\usepackage{caption}
\usepackage{graphicx}
\usepackage{multirow}
\usepackage{url}
\usepackage{makecell}

\usepackage[style=numeric]{biblatex}
\title{Large Language Models\\
       for Biomedical Article Classification}
\author{Jakub Proboszcz}
\author{Paweł Cichosz}
\affil{Institute of Computer Science\\
       Warsaw University of Technology\\
       Nowowiejska 15/19, 00-665 Warsaw, Poland\\
       \{jakub.proboszcz.stud, pawel.cichosz\}@pw.edu.pl}
\date{}

\begin{document}
\sloppy

\maketitle
\begin{abstract}
This work presents a systematic and in-depth investigation of the utility of large language models as text classifiers for biomedical article classification. The study uses several small and mid-size open source models, as well as selected closed source ones, and is more comprehensive than most prior work with respect to the scope of evaluated configurations: different types of prompts, output processing methods for generating both class and class probability predictions, as well as few-shot example counts and selection methods. The performance of the most successful configurations is compared to that of conventional classification algorithms. The obtained average PR AUC over $15$ challenging datasets above $0.4$ for zero-shot prompting and nearly $0.5$ for few-shot prompting comes close to that of the na\"ive Bayes classifier ($0.5$), the random forest algorithm ($0.5$ with default settings or $0.55$ with hyperparameter tuning) and fine-tuned transformer models ($0.5$). These results confirm the utility of large language models as text classifiers for non-trivial domains and provide practical recommendations of the most promising setups, including in particular using output token probabilities for class probability prediction.
\end{abstract}

\section{Introduction}

In recent years, the development of deep learning applied to natural language processing has given rise to large language models (LLM) -- neural networks based on the transformers architecture~\cite{vaswani2023attentionneed}, trained on huge text corpora to predict text continuation and often fine-tuned to chat or follow user instruction. Due to their emergent abilities, these models can be successfully used for tasks not only involving text generation, but also text understanding and reasoning~\cite{wei2022emergentabilitieslargelanguage}, without additional finetuning. These tasks include machine translation~\cite{llm_translation2023}, text summarization~\cite{zhang2025summarization}, code generation~\cite{llm_coding2021}, question answering~\cite{tan2023llmqa}, mathematical reasoning~\cite{imani2023mathprompter}, or text classification~\cite{sun23classification,chae23classification,wang23zeroshot,edwards24textclassification}. Using a large language model for any of such tasks requires designing an appropriate prompt, which may or may not include some examples of inputs and desired outputs. Prompts with one or more examples are referred to as one- or few-shot prompts whereas prompts with no examples are referred to as zero-shot prompts.

\subsection{Motivation}

Text classification is one of the most commonly studied tasks at the intersection of natural language processing and machine learning \cite{mccallum98event,joachims98categorization,dumais98categorization,sebastiani02categorization,aggarwal12text,wu14forestexter,wang20comparative,cichosz23repclas}. The standard approach is to train a classification model from a training set of documents with class labels, transformed to a vector representation. It often yields satisfactory classification quality but is hardly applicable when labeled training documents are scarce or not available at all. This is the case, in particular, when class labels are obtained during a human annotation process, which is time consuming and costly. This obstacle is only partially alleviated by approaches reducing the amount of required labeled data, such as active learning~\cite{tong01svm,jacobs22reducing,cichosz24repal} or self-training \cite{schroder24selftraining}.

The primary appeal of LLMs as text classifiers is that they do not need labeled training data at all when used with zero-shot prompts, and with few-shot prompts they only need a very small number of labeled examples. What they require instead is that class labels are interpretable or class membership criteria can be specified in natural language. In several application domains these requirements are easier to satisfy than those of conventional machine learning algorithms. Even if labeled data for classification model creation is available or can be collected, LLMs can sometimes be an attractive alternative whenever ``understanding'' text contents and classes has more utility than generalizing observed relationship patterns between text representations and classes. This is likely to be the case for specialized domains that require some background knowledge as well as deeper text comprehension to properly assign classes. One such domain may be that of biomedical articles, where texts of all classes usually belong to the same thematic area and it may be hard to discern classes based just on word occurrence statistics. It is possible that large language models could utilize the knowledge encoded in billions of their parameters to achieve results as good as or better than conventional classifiers trained on labeled data.

\subsection{Related Work}

With the onset of large language models, their increasing capabilities and accessibility stimulate a growing stream of publications reporting their use for text classification. Table~\ref{tab:related.work} summarizes recent work that is the most closely related to the scope and aims of this article along the following dimensions:
\begin{description}
\item[data:] text datasets or classification tasks,
\item[models:] LLMs used for text classification,
\item[prompts:] types of prompts or prompt design techniques,
\item[example selection:] methods used to select examples for few-shot prompts,
\item[evaluation:] performance measures and evaluation procedures.
\end{description}
Prompts are categorized as zero- or few-shot (one-shot in some cases) and their complexity is characterized. The use of prompt design techniques such as chain of thought (CoT)~\cite{wei2023chainofthoughtpromptingelicitsreasoning} or clue and reasoning prompting (CARP)~\cite{sun23classification} is also indicated where appropriate. When semantic similarity was used for example selection, the corresponding embedding model is provided. Whenever one- or few-shot prompts were used, the procedure used to separate training and test documents is specified in addition to classification quality measures in the evaluation column.

An interesting aspect of study design is omitted in the table: the type of model output or output processing used. This is because the prior studies do not differ in output processing methods aside from minor variations, the most significant of which is using the JSON format or plain text for model answers. Nevertheless, this aspect is important to explain the novelty of this work and is therefore addressed in the discussion presented below.

{\footnotesize\hyphenpenalty=100
  \begin{longtable}{>{\raggedright}p{0.04\textwidth}|>{\raggedright}p{0.15\textwidth}|>{\raggedright}p{0.18\textwidth}|>{\raggedright}p{0.13\textwidth}|>{\raggedright}p{0.18\textwidth}|>{\raggedright\arraybackslash}p{0.12\textwidth}}
  \caption{Related work.}
  \label{tab:related.work}\\
  \textbf{Ref.} & \textbf{Data} & \textbf{Models} & \textbf{Prompts} &
  \textbf{Example Selection} & \textbf{Evaluation}\\
  \hline
  \cite{sun23classification}
  & sentiment phrases, movie reviews, news articles
  & GPT-3 DaVinci
  & zero-shot, few-shot; vanilla, CoT, CARP
  & random, SimCSE~\cite{gao2022simcsesimplecontrastivelearning} and fine-tuned SimCSE embedding
  & accuracy; train-test split\\
  \hline
  \cite{wang23zeroshot}
  & COVID-19 tweets, economic and e-commerce texts, SMS spam detection
  & Llama-2-70B-Instruct, GPT-3.5 Turbo, GPT-4
  & zero-shot; vanilla
  & -
  & accuracy, F1 \\
  \hline
  \cite{chae23classification}
  & tweets and Facebook comments on US elections
  & FLAN-T5 XXL, Mistral-7B, Llama-3-8B, Llama-3-70B, GPT-3 Ada, GPT-3 DaVinci, GPT-4o (all except GPT-4o fine-tuned)
  & zero-shot and one/few-shot (one-shot for threads); varying complexity
  & random, with class balancing
  & F1; bootstrapping\\
  \hline
  \cite{wang25adaptable}
  & COVID-19 tweets, economic and e-commerce texts, SMS spam detection
  & GPT-3.5 Turbo, GPT-4, Gemini-pro, Llama-3-8B Instruct (fine-tuned), Qwen-Chat 7B (fine-tuned), Qwen-Chat 14B, Vicuna v1.5 7B and 13B
  & zero-shot, few-shot; vanilla
  & manual
  & accuracy, F1, uncertainty-error rate (custom metric of LLM's refusals to answer); train-test split\\
  \hline
  \cite{edwards24textclassification}
  & tweets, news articles, movie reviews, Wikipedia comments, medical abstracts, legislation documents, safeguarding reports
  & Llama-1-7B, Llama-2-7B, Flan-T5 780M, GPT-3.5 Turbo
  & zero-shot, few-shot with one example per class; varying scope of task/domain description
  & random, with class balancing
  & F1; train-test split\\
  \hline
  \cite{guo24evaluating}
  & healthcare-related tweets
  & GPT-3.5, GPT-4
  & zero-shot; vanilla
  & -
  & precision, recall, F1; k-CV\\
  \hline
  \cite{labrak24zeroshot}
  & medical article abstracts, COVID-19 literature, public health claims, discharge summaries
  & GPT-3.5 Turbo, FLAN-T5 UL2, Tk-Instruct, Alpaca
  & zero-shot, few-shot; vanilla
  & PubMedBERT \cite{gu2021pubmedbert} embedding
  & accuracy, F1; train-test split\\
  \hline
  \cite{rathje24gpt}
  & tweets, Reddit comments, news headlines (sentiment analysis, offensiveness detection, emotion detection)
  & GPT-3.5 Turbo, GPT-4, GPT-4 Turbo
  & zero-shot; vanilla
  & -
  & accuracy, F1\\
  \hline
  \cite{chen24evaluating}
  & sentences from medical articles
  & GPT-3.5 Turbo, GPT-3.5 DaVinci, GPT-4, GPT-J, GPT-JT, Galactica
  & zero-shot, one/few-shot; vanilla, CoT
  & manual
  & F1; k-CV\\
  \hline
  \cite{zhang2025pushingthelimit}
  & medical article references, movie reviews, news articles
  & ensemble of fine-tuned Llama 2 models
  & zero-shot, one-shot, few-shot; vanilla
  & unspecified
  & accuracy, F1; train-test split\\
  \hline
  \cite{singh2025optimizingpromptrefinement}
  &  technical certification exams questions
  & GPT-4.0
  & zero-shot, few-shot; vanilla, CoT, self-consistency, ToT, ranking different reasonings, varying classification criteria and combining outputs
  & manual, keyword search, maximal marginal relevance, Ada-002~\cite{openai2025embeddings} and all-MiniLM-L6-v2 embedding, hybrid with reranking
  & precision, recall, F1; train-test split\\
  \hline
  \cite{shi2026anempiricalstudyofllms}
  & tweets, blog and forum comments, news and encyclopedia articles
  & Qwen 2.5 7B Instruct, Qwen 2.5 72B Instruct, DeepSeek-R1-Distill-Qwen-7B, DeepSeek-R1-Distill-Llama-70B, Llama 3.1 8B Instruct, Llama 3.1 70B Instruct, GPT-3.5, Gemini-1.5
  & zero-shot, few-shot; vanilla
  & random, Qwen 3 embedding~\cite{aliyun2025textembeddingv3}
  & F1; train-test split 
  \end{longtable}
}

At the first glance, these prior studies may appear to cover nearly all interesting research issues, leaving little space for novel contributions: they include both commercial and open source LLMs of varying size, prompts of varying complexity, and datasets from different application domains. On a closer look, however, it can be seen that many of them use only basic prompt designs, address classification tasks that are not particularly demanding, and limit performance evaluation to the most basic classification quality measures such as accuracy or F1, estimated using the simple train-test split procedure (in the case of methods to which training and test data separation is applicable).  

While it is not visible in Table~\ref{tab:related.work}, due to the lack of a corresponding column, all of prior studies listed there instruct the LLM to respond by providing class labels and directly use its textual output to determine class predictions, making no attempt to obtain any measure of prediction confidence (e.g., by instructing the model to output such a value or by processing output token probabilities). This is why only basic classification quality measures based on comparing true and predicted class labels are applicable. More informative measures, such as the area under the ROC or precision-recall curve, cannot be therefore calculated since they require some kind of numeric score as prediction output (in the case of standard classifiers, these are predicted class probabilities or decision function values).  This is not just a limitation of evaluation, but also a limitation of usability: there is no way of adjusting a decision threshold to compensate for class imbalance or increase sensitivity to more important classes with higher misclassification costs. This may be a severe drawback in some application domains, particularly when certain class predictions may trigger actions that require the allocation of limited resources.

\subsection{Objectives}

The objective of this work is to at least partially fill gaps left by prior research by more deeply and systematically examining the utility of large language models as text classifiers for the demanding domain of medical scientific article classification and identifying the most useful setups in which they can be applied. More specifically, this includes verifying how well large language models perform in classification tasks with severe class imbalance and vague class boundaries, how this performance is impacted by prompt designs, output processing methods, as well as few-shot example counts and selection criteria, and how it compares to that of conventionally trained classification models.

\subsection{Contributions}

To achieve the research objectives formulated above, this article  reports the results of experimental studies which are more extensive and in-depth than those previously presented in the literature. While the scope of LLMs used remains limited due to computational resource and budget limitations, the scope of prompt designs is more complete and classification quality evaluation is more thorough than in most of previously published experiments. Moreover, different model output processing methods to obtain prediction confidence and example selection methods for few-shot LLM classification are experimented with. More specifically, the contributions of this work can be summarized as follows.
\begin{itemize}
\item Evaluating the utility of LLMs as text classifiers for biomedical scientific articles with respect to inclusion or exclusion in systematic literature reviews for $15$ drug class-related topics.
\item Using several openly available LLMs: Llama 3.1 8B Instruct, Llama 3.1 70B Instruct, Mistral 7B v0.3 Instruct and Gemma 2 9B Chat, with  OpenAI GPT-4.1 mini and GPT-4.1 nano models additionally included in the final comparison study.
\item Exploring the impact of prompts of varying complexity on the quality of classification.
\item Experimenting with several methods of parsing model output and obtaining prediction confidence: prompting the model to return an answer in JSON format, prompting the model to give a numeric score for the sample, calculating class probabilities based on token probabilities.
\item Applying different example selection methods for few-shot learning: random selection, semantic similarity-based selection using different embedding models (all-MiniLM-L6-v2, all-distilroberta-v1, all-mpnet-base-v2), similarity-based selection with class rebalancing.
\item Evaluating advanced prompting methods: chain of thought with single and double model requests, tree of thoughts, and aggregating model answers for chunks of the classified sample.
\end{itemize}

To draw reliable conclusions, performance is evaluated using several classification quality measures for both discrete class predictions (precision, recall, F1, Matthew's correlation coefficient) and numeric score or probability predictions (area under the precision-recall curve), using $k$-fold cross-validation for few-shot prompts, and the identified best LLM configurations are compared to conventional machine learning classifiers, na\"{i}ve Bayes and random forest with the bag of words text representation, as well as to fine-tuned transformer encoder models: DeBERTa and SciDeBERTa-v2.

\section{LLMs as Text Classifiers}

While large language models can be used to perform the same text classification task that has been addressed using conventional machine learning algorithms for years, it might seem like taking a sledgehammer to crack a nut, given the impressive text generation, question answering, and reasoning capabilities of LLMs as well as their extremely high computational demands. Contrary to this superficial judgment, using LLMs as text classifiers is quite well justified since, unlike conventional machine learning algorithms, they can classify texts using no or very few training examples, based only or mostly on meaningful class labels or provided class membership criteria. This does not make conventional text classifiers obsolete by any means, though, since whenever sufficient labeled training data is available, they can often provide high-quality class predictions using a tiny fraction of computational resources needed for LLMs. 

\subsection{Text Classification}

Text classification is the task of mapping text samples to discrete classes. This could be separating spam from ham in email messages, identifying positive or negative sentiment, assigning texts to topic categories, etc. Several conventional machine learning algorithms are commonly used for text classification, including multinomial na\"{\i}ve Bayes classifier \cite{mccallum98event} or random forest \cite{breiman01rf}.

Such algorithms, originally designed for tabular data, require the text to be represented as a vector of attributes. A typical method of text representation is bag of words, where the $i$-th attribute is the count of how many times the $i$-th word in the vocabulary appeared in the text sample, which is referred to as term frequency. This representation can be modified using TF-IDF, which normalizes term frequency by inverse document frequency, based on the number of documents in the dataset that contain a given word \cite{aggarwal12text}. Another possible extension is to use $n$-grams (word sequences of length~$n$) as terms rather than just single words. More advanced embedding-based text representations have also become popular \cite{mikolov13efficient,pennington14glove,le14documents,bojanowski2016enriching,akbik2018contextual}.

A necessary condition for applying conventional machine learning algorithms is the availability of a sufficiently large labeled dataset, containing texts with true classes assigned. These serve as training examples for classification model creation. In several interesting application domains this requirement is difficult or costly to satisfy, since training class labels need to be assigned by human evaluators.

Neural networks based on encoder-only transformer architecture~\cite{vaswani2023attentionneed} have also been used for text classification. These models are usually first pre-trained using the masked language modeling objective. There are numerous such models available online, including BERT~\cite{devlin19bert}, RoBERTa~\cite{liu2019roberta} and DeBERTa~\cite{he2021deberta}. They can later be fine-tuned to the specific classification task after adding a classification head -- one or several neural network layers that output class probabilities.

\subsection{Large Language Models}

Large language models are deep neural networks based on the transformer architecture~\cite{vaswani2023attentionneed}. Most of them utilize a decoder-only version thereof, with further modifications.

Training an LLM is usually split into at least two stages:
\begin{itemize}
    \item pre-training with the objective of next word prediction on large text corpora,
    \item fine-tuning to improve the LLM's ability to chat or follow instructions and complete tasks like question answering or coding.
\end{itemize}

Both pre-trained-only base text-completion models and fine-tuned models can be used for various tasks. Applying either variant of LLMs requires the preparation of a prompt -- the input text for the model. In addition to that, a method of processing the model's output (either the generated text or, if available, raw token probabilities) into answers appropriate for the given task is needed.

For this article's experiments, six large language models were selected. Most experiments were done on four openly available LLMs: Llama 3.1 8B, Llama 3.1 70B~\cite{grattafiori2024llama3herdmodels}, Gemma 2 9B~\cite{gemmateam2024gemma2improvingopen}, Mistral 7B~\cite{mistral7b2023}. Additionally, in the final comparison two closed-source models were included: GPT-4.1 mini and GPT-4.1 nano~\cite{openai2025models}. Openly available models trained by various authors were selected, with relatively small parameters counts in order to limit the computing power necessary to run the experiments. Experiments on the closed-source, paid models were limited because of the cost.

\subsection{Prompts}

A prompt is the input of a large language model. Preparing a prompt is necessary to use an LLM for any purpose. Generally, a good prompt should clearly explain the problem to be solved, provide the necessary context and specify the output format. A prompt may contain examples of correct solutions to problems similar to the one that should be solved by the LLM. This is called a few-shot prompt, while a prompt without such examples is called a zero-shot prompt. In the case of text classification, the examples are selected from a subset of text samples for which true class labels are available.

In the case of base text-completion models, the prompt has to be composed in such a way that its continuation generated by the model contains the expected answer. This is not necessary for models fine-tuned for conversation and to follow the user's instruction (so called chat or instruct models) which can answer directly asked questions or generate output as instructed in the prompt.

For chat and instruction fine-tuned models there are certain model-specific prompt templates, with special tokens used to mark different parts of the prompt. These may include so called system and user messages, the former usually providing a general description of the task and requirements for the answer and the latter usually providing a specific input for which the model is required to reply. For text classification, the system message could describe the classification task (the target concept), possible classes, and specify the expected answer format, whereas the user message could present a specific text sample to classifier.

Prompts may also contain assistant messages, corresponding to LLM answers from a real or hypothetical interaction with the user. In the case of chat applications they are used to represent conversation history, but for non-chat applications they may also be used to provide examples for few-shot prompts. For text classification, a series of user messages and the corresponding assistant messages would present text samples and the corresponding true classes.

\subsection{Example Selection}

When using few-shot prompts for text classification, a set of text samples is assumed to be available for which true class labels are known or can be obtained. These will be referred to as training samples, although strictly speaking no actual training is taking place, since they are only used as examples demonstrating correct classification to the model.

There are several possible methods of example selection for few-shot prompts. The simplest method is to select examples manually. These examples are then a part of the prompt independent of the classified sample. It is also possible to select the examples randomly from the training set. However, selecting examples based on semantic similarity is thought to give better results.

For similarity-based selection, all the training samples have their embeddings calculated using a model independent from the LLM used for classification. Then, for each classified sample, its embedding is calculated and examples are selected from the training set using the $k$ nearest neighbors algorithm. The distance measure for neighbors selection is usually L2 or cosine distance. In the case of embedding vectors which are normalized to unit length, which is quite common for popular embedding models, these two measures are equivalent.

In the case of an unbalanced dataset, it might be a common occurrence that all selected examples have the same class. The LLM might then answer with this class following the examples and ignoring the actual classified sample. It is possible to prevent this situation by enforcing 
a more balanced representation of all classes among the selected examples, e.g., specifying the number of training samples to select from particular classes.

\subsection{Output Processing}

The output of an LLM is the text generated after feeding the model with the prompt, that is, a sequence of tokens. To convert it to a class, it is necessary to associate each class with a word or phrase. The required model answer format has to be specified in the prompt, so that the model actually generates answers in the expected form.

The simplest method of extracting the class from the output text is to search for the expected words or phrases -- encountering a phrase means that the associated class is returned. Typically, the model would be instructed to provide just the class label on output. An alternative approach is to request a structured answer format such as a JSON object with a single field, where the key would represent the target concept name and the value would represent the class label. 

While these methods are sufficient for text classification, they do not provide class probabilities or any other scores that could be used to adjust the decision threshold, which is a highly desirable capability in many applications. One possible way to obtain them could be to request the model in the prompt to rate how likely the classified sample is to belong to particular classes, or -- in the case of binary classification -- to the positive class. The answers then contain numbers, which after normalization can be considered probabilities of the positive class.

Another way to obtain class probabilities is to calculate them based on the probability distribution of the first token generated by the model. In this approach, the probabilities of tokens corresponding to particular class labels, normalized to add to $1$, would be considered predicted class probabilities (if there are multiple possible tokens corresponding to the same class, e.g., with different spelling, their probabilities would be summed up).

\subsection{Advanced Techniques}

Some techniques known from the literature for more complex LLM applications  may be also useful for text classification. These include chain of thought, tree of thoughts, and chunking methods.

\subsubsection{Chain of Thought}

Chain of thought~\cite{wei2023chainofthoughtpromptingelicitsreasoning} is a prompt engineering technique in which the LLM is prompted to generate a reasoning before giving the final answer. This may be as simple as adding an appropriate phrase to the prompt (such as "think step by step") or some more detailed instruction about specific reasoning steps to perform before producing the final answer. It may also involve chaining two or more model inference requests, with the first and intermediate ones generating some reasoning results and the final one wrapping them up into the final answer. In the case of text classification, an intermediate step could be a list of phrases from the classified sample that may be relevant to the classification task.

\subsubsection{Tree of Thoughts}

Tree of thoughts~\cite{yao2023treethoughtsdeliberateproblem} is a generalization of chain of thought with multiple LLM requests, referred to as steps. Each step is a piece of reasoning generated via a single LLM request, like in the chain of thought method. In this method, after each step an additional LLM request is made to score the obtained answer. Afterwards, if the score is below a certain threshold, the answer is rejected and the step is repeated. If the score is equal or above the threshold, the answer is accepted and the next step begins, using the accepted answer as input. If an answer for the same step is rejected more than a specified number of times, the algorithm goes back to the previous step and repeats it as if its answer was rejected. If this happens for the first step, the tree of thoughts process fails (does not provide a valid answer).

\subsubsection{Chunking}

Chunking is a method of splitting long documents before inserting them into LLM input, typically used to handle input contents that do not fit in the model's context length limit. It could be useful for text classification even if the sample to be classified is well below the context limit. The idea is to split it into chunks anyway, use separate LLM requests to classify each of these chunks, and combine the answers. This is motivated by the hypothesis that working on shorter text portions and aggregating results may improve classification performance on complex tasks, with nuanced class membership criteria.

\subsection{Implementation}

The experiments were conducted using a custom command-line tool implemented in Python 3.11. The following libraries were used:
\begin{itemize}
    \item langchain~\cite{chase2025langchain} version 0.3.4 for a simple unified LLM interface,
    \item llama-cpp-python~\cite{llamacpppython2025docs} version 0.3.16 for handling of locally-run LLMs,
    \item scikit-learn~\cite{scikit-learn} version 1.5.2 for conventional machine learning algorithms and classification quality evaluation,
    \item HuggingFace transformers~\cite{huggingfacetransformers} version 4.57.6 for fine-tuning baseline transformer models,
    \item sentence-transformers~\cite{reimers2019sentencebertsentenceembeddingsusing} version 3.2.0 for embedding models.
\end{itemize}

\section{Experimental Studies}

To assess the utility of large language models in several different setups for non-trivial text classification tasks, they were applied to datasets from biomedical systematic literature reviews.

\subsection{Datasets}

Experiments were conducted on a set of biomedical articles knows as Drug Class Review (DCR) data~\cite{dcr15}, containing PubMed article identifiers with the corresponding expert inclusion or exclusion decisions in systematic literature review studies on 15 topics, related to different drug groups. The expert first read the article's abstract and decided whether to immediately reject the article. If the article was not rejected, the expert read the entire publication and made a final decision about its inclusion in the systematic review.

To use the DCR data for text classification, article abstracts were fetched from the PubMed database and merged with the classes corresponding to expert decisions made at the abstract review stage for each of the 15 study topics. This yields datasets for 15 binary classification tasks, where the classifier's input is the article abstract and the expected output is the inclusion or exclusion decision on that abstract.

Table~\ref{tab:dcr.stats} contains the most important statistics about each of the 15 datasets.

\begin{table}[!htb]
\caption{Statistics of DCR datasets}\label{tab:dcr.stats}
\resizebox{\linewidth}{!}{%
\begin{tabular}{|c|c|c|c|}
\hline
Topic & \# of samples & Positive class \% & Avg. \# of words in sample\\
\hline
\hline
ACEInhibitors & $2236$ & $7.6$ & $243.9\pm 84.5$\\
\hline
ADHD & $803$ & $10.3$ & $222.9\pm 78.6$\\
\hline
Antihistamines & $287$ & $31.4$ & $224.1\pm 79.3$\\
\hline
AtypicalAntipsychotics & $1031$ & $32.3$ & $211.5\pm 81.0$\\
\hline
BetaBlockers & $1876$ & $14.4$ & $240.4\pm 84.7$\\
\hline
CalciumChannelBlockers & $1109$ & $23.2$ & $244.7\pm 80.6$\\
\hline
Estrogens & $349$ & $22.6$ & $272.0\pm 84.0$\\
\hline
NSAIDS & $358$ & $23.2$ & $232.2\pm 85.2$\\
\hline
Opiods & $1772$ & $2.4$ & $220.8\pm 89.4$\\
\hline
OralHypoglycemics & $475$ & $28.4$ & $252.4\pm 89.3$\\
\hline
ProtonPumpInhibitors & $1211$ & $18.7$ & $240.8\pm 72.4$\\
\hline
SkeletalMuscleRelaxants & $1353$ & $2.2$ & $181.5\pm 92.0$\\
\hline
Statins & $2745$ & $5.5$ & $226.1\pm 89.9$\\
\hline
Triptans & $594$ & $34.5$ & $223.0\pm 95.8$\\
\hline
UrinaryIncontinence & $284$ & $23.9$ & $224.7\pm 80.8$\\
\hline
\end{tabular}
}
\end{table}

\subsection{Models}

Four open source large language models are evaluated in this study: Gemma 2 9B Instruct~\cite{gemma29b_quant}, Llama 3.1 8B Instruct~\cite{llama3.18b_quant}, Llama 3.1 70B Instruct~\cite{llama3.170b_quant} and Mistral 7B v0.3 Instruct~\cite{mistral7b_quant}. These models were chosen because they are some of the most commonly used models with weights available to download and run locally. Due to the limited computing power available, only small and medium-size models were selected: three with less than~10 billion parameters and one with 70~billion parameters.

Quantized versions of models were utilized in order to speed up computations. The chosen quantization variant was Q4\_K\_M. The quantized models were downloaded from HuggingFace repositories.

Additionally, in the final comparison experiment, two closed source OpenAI models are used, GPT-4.1 mini and GPT-4.1 nano. They are mid-size and small representatives of the last (at the time of this writing) series of models for which token probabilities are available through OpenAI API.

As the training datasets of any of the used LLMs are not known and the DCR datasets predate the creation of these models, it is possible that the evaluation data for this article's experiments was part of the models' training sets. However, because the open-source LLMs used were run locally and OpenAI declares that data sent via its API is not used for training~\cite{openai2025dataprivacy}, these experiments have not contributed to any further overfitting of future models on evaluation datasets.

\subsection{Performance Evaluation}

The performance of the classifiers was evaluated using the following metrics:
\begin{itemize}
    \item accuracy,
    \item recall,
    \item precision,
    \item F1 score,
    \item Matthews' correlation coefficient (MCC),
    \item area under the precision-recall curve (AUPRC).
\end{itemize}

All these metrics were calculated via stratified 5-fold cross-validation. For each fold the training subset was used to select examples for few-shot prompts or and ignored for zero-shot prompts.

In tables showing MCC or AUPRC scores of each configuration per dataset, the results are macro-averaged over the cross-validation iterations and over the four LLMs experimented with. An exception is the final experiment in which the best configurations are compared with conventional classification algorithms, where there is no aggregation over models and the results for each LLM are given separately, macro-averaged only over cross-validation iterations. In these tables, for each dataset the best score is marked in bold. The configuration with the highest score macro-averaged over all datasets also has its header bolded.

In tables with performance scores aggregated over all models, the results are macro-averaged over cross-validation iterations and over LLMs. In tables with performance scores aggregated over all datasets, the results are macro-averaged over cross-validation iterations and over datasets. The rows marked as "Average" are also macro-averaged over LLMs. For each score, the best result among different configurations for each particular model, as well as the best score averaged over all models are marked in bold. The configuration with the highest AUPRC score (or MCC in the case of the first experiment in which class probabilities were not estimated) macro-averaged over all models also has its header marked in bold.

For comparison between configurations returning class probabilities and those returning only the predicted classes, in order to calculate AUPRC, the results of the latter variants were assigned probabilities based on the predicted class:
\begin{itemize}
    \item 0.0 if the negative class was predicted,
    \item 1.0 if the positive class was predicted,
    \item 0.5 if there was an unrecognized answer.
\end{itemize}
The last possibility is included for completeness, to handle those rare cases where the model did not follow the instruction specifying the required answer format in the prompt.

For configurations returning class probabilities, the positive class was assigned to the sample if the returned probability was greater than 0.5.

In order to give insight into the performance of the studied classifiers at various decision thresholds, Appendix~\ref{app:prc} also contains precision-recall curves for the configurations where probabilities are available. For each configuration, the precision-recall curve of the best model with this configuration (selected using macro-averaged AUPRC score) is displayed. The AUPRC values presented in plot legends are the areas under the corresponding curves, obtained by combining true class labels and predictions from all cross-validation folds, i.e., by micro-averaging, unlike the results in the tables which are macro-averaged. This is because averaging with interpolation does not provide reliable results for precision-recall curves \cite{davis06prroc}. These curves are presented to illustrate the scope of precision-recall tradeoffs corresponding to different decision thresholds, but the best configuration to use in the next experiment is always selected based on the macro-averaged performance scores in the tables.

\subsection{Experimental Studies}

Subsequent subsections contain experimental studies exploring the following areas.
\begin{description}
    \item[Prompts:] prompts of varying complexity and user message contents.
    \item[Output:] output processing methods: answer as a single word, answer in JSON format, answer as a number representing positive class probability, calculating class probabilities based on first token probabilities.
    \item[Few-Shot-Count:] the number of examples used for few-shot prompts.
    \item[Few-Shot-Selection:] example selection methods for few-shot prompts: random selection, selection based on semantic similarity in embedding space of three different embedding models, semantic similarity selection of examples with balanced classes.
    \item[Advanced-Techniques:] chunking, chain of thought with single or double model requests, tree of thoughts.
    \item[Baseline:] comparison with baseline classification algorithms: multinomial na\"{i}ve Bayes, random forest and transformer encoder models, as well as with closed source OpenAI models.
\end{description}

\subsection{Prompts Experiment}

The first experiment explores the impact of prompt choice on classification quality. Six prompts were prepared, varying in both complexity and the contents of system and user messages. Table~\ref{tab:dcr.prompts_chars} presents the properties of each of the prompts used. Simple prompts contain at most two short sentences per message, medium prompts contain no more than three sentences and complex prompts are longer than that. Expected model role contained in a prompt is the information that the model is supposed to simulate an expert in medical articles classification. Information about task specifies the objective of the classification task -- a general description of the target concept (inclusion or exclusion in a systematic literature review on a given topic), as well as the expected answers (yes or no), while question about task is a direct specific question whether the given abstract should be included in a review on the given topic. Appendix~\ref{app:prompts} contains the full prompts used in all experiments.

The prompts instruct the model to decide whether the article is high-quality and related to the topic of the review. Unlike in the most typical approach to classification for systematic literature reviews, which only classifies whether the article is relevant to the topic, in the Drug Class Review datasets articles were rejected due to inadequate quality, i.e., failure to meet certain criteria (among rejection reasons were: ``wrong outcome'', ``wrong population'', ``wrong study design''). For this reason, a mention of the quality of the article was added to the prompts. Preliminary experiments confirmed a classification performance improvement after this change.

All models used in this work are instruction fine-tuned models. Two of those (Mistral 7B and Gemma 2) do not support system messages, so the task has to be entirely defined in the user message. For these two models, the intended contents of the system message is prepended to the user message, separated with two newline characters. The other two, from the Llama 3 family, support system messages, so the classification task specification and specific classification requests can be split among system and user messages. 

In this initial experiment, the zero-shot setup and basic output processing method were used: the prompts instruct the LLM to output the predicted class directly. The expected answers were \texttt{no} and \texttt{yes}, associated with the negative and positive class respectively. The expected answers were used as stop sequences, i.e., after detecting any of them generation was interrupted, so there was no possibility of encountering both of them in the model output. If neither answer was detected, the negative class was returned, as it is the majority class in all 15 datasets. Model output was limited to 256 tokens -- a large limit so as to avoid possible infinite, repeating generations but not to interrupt long but potentially partially correct answers. Deterministic sampling of generated tokens was utilized by setting the temperature parameter to zero.

\begin{table}[!htb]
\caption{Prompts used in the Prompts  experiment}\label{tab:dcr.prompts_chars}
\centering
\begin{tabular}{|c|c|c|c|}
\hline
Prompt & Complexity & System message & User message\\
\hline
\hline
A & simple & question about task & only article abstract\\
\hline
\multirow{2}{*}{B} & \multirow{2}{*}{simple} & \multirow{2}{*}{expected model role} & \multirow{2}{*}{\shortstack{question about task \\ article abstract}}\\
& & & \\
\hline
\multirow{2}{*}{C} & \multirow{2}{*}{medium} & \multirow{2}{*}{\shortstack{expected model role \\ information about task}} & \multirow{2}{*}{only article abstract}\\
& & & \\
\hline
\multirow{2}{*}{D} & \multirow{2}{*}{medium} & \multirow{2}{*}{\shortstack{expected model role \\ information about task}} & \multirow{2}{*}{\shortstack{question about task \\ article abstract}}\\
& & & \\
\hline
\multirow{2}{*}{E} & \multirow{2}{*}{complex} & \multirow{2}{*}{\shortstack{expected model role \\ information about task}} & \multirow{2}{*}{only article abstract}\\
& & & \\
\hline
\multirow{2}{*}{F} & \multirow{2}{*}{complex} & \multirow{2}{*}{\shortstack{expected model role \\ information about task}} & \multirow{2}{*}{\shortstack{question about task \\ article abstract}}\\
& & & \\
\hline
\end{tabular}
\end{table}

Table~\ref{tab:dcr.prompt.mcc} contains the macro-averaged MCC scores for each of the prompts. In addition to that, Table~\ref{tab:dcr.prompt.avg} in Appendix~\ref{app:models} contains more performance scores for each large language model averaged over all the datasets. The two best prompts according to the average MCC score: B and F were selected for the next experiment. There appears to be no clear correlation between prompt complexity and classification quality -- of the two prompts with the highest average MCC score one is simple and one is complex. However, the strategy of adding a question describing the task to the user message has proven effective: prompts B, D, F achieved equal or better MCC scores than their counterparts of equivalent complexity for all datasets except UrinaryIncontinence.

The number of unrecognized answers was negligible -- it did not exceed $0.03$\% for any of the tested dataset-prompt pairs. In these rare cases, the LLM did not conform to the required answer format specified in the prompt and output some kind of a reasoning that was cut off by the output token limit, or gave an uncertain answer, for example:
\begin{verbatim}
This abstract seems relevant to BetaBlockers as they are often
used in the treatment of migraines.
**However, it doesn't explicitly mention BetaBlockers.**

To confidently classify it, I'd need to see the full article.

**I'd say: Maybe.**
\end{verbatim}

\begin{table}[!htb]
\caption{Macro-averaged MCC scores for the Prompts experiment.}\label{tab:dcr.prompt.mcc}
\resizebox{\linewidth}{!}{%
\begin{tabular}{|c|c|c|c|c|c|c|}
\hline
Prompt & A & B & C & D & E & \textbf{F} \\
\hline\hline
ACEInhibitors & $0.080$ & $\mathbf{0.106}$ & $0.080$ & $0.094$ & $0.077$ & $0.091$ \\
\hline
ADHD & $0.402$ & $0.441$ & $0.436$ & $0.476$ & $0.490$ & $\mathbf{0.523}$ \\
\hline
Antihistamines & $0.163$ & $0.214$ & $0.175$ & $0.183$ & $0.187$ & $\mathbf{0.254}$ \\
\hline
AtypicalAntipsychotics & $0.093$ & $0.143$ & $0.086$ & $0.122$ & $0.116$ & $\mathbf{0.147}$ \\
\hline
BetaBlockers & $0.093$ & $0.129$ & $0.110$ & $0.126$ & $0.113$ & $\mathbf{0.130}$ \\
\hline
CalciumChannelBlockers & $0.094$ & $0.119$ & $0.100$ & $0.113$ & $0.104$ & $\mathbf{0.126}$ \\
\hline
Estrogens & $0.154$ & $\mathbf{0.183}$ & $0.126$ & $0.170$ & $0.136$ & $0.173$ \\
\hline
NSAIDS & $0.090$ & $\mathbf{0.115}$ & $0.072$ & $0.101$ & $0.079$ & $0.096$ \\
\hline
Opiods & $0.034$ & $0.050$ & $0.049$ & $0.049$ & $0.047$ & $\mathbf{0.059}$ \\
\hline
OralHypoglycemics & $0.118$ & $0.132$ & $0.066$ & $0.094$ & $0.087$ & $\mathbf{0.137}$ \\
\hline
ProtonPumpInhibitors & $0.082$ & $\mathbf{0.135}$ & $0.097$ & $0.120$ & $0.098$ & $0.115$ \\
\hline
SkeletalMuscleRelaxants & $0.150$ & $\mathbf{0.185}$ & $0.138$ & $0.170$ & $0.142$ & $0.155$ \\
\hline
Statins & $0.088$ & $0.104$ & $0.096$ & $0.105$ & $0.092$ & $\mathbf{0.111}$ \\
\hline
Triptans & $0.246$ & $0.291$ & $0.244$ & $0.285$ & $0.262$ & $\mathbf{0.299}$ \\
\hline
UrinaryIncontinence & $0.083$ & $\mathbf{0.102}$ & $0.097$ & $0.083$ & $0.062$ & $0.042$ \\
\hline
\hline
\textbf{Average} & $0.131$ & $0.163$ & $0.131$ & $0.153$ & $0.139$ & $\mathbf{0.164}$ \\
\hline
\end{tabular}
}
\end{table}

\subsection{Output Experiment}

In this experiment, the following methods of processing LLM output were compared.
\begin{description}
    \item[Word:] answer as a single word (base method, used in the previous experiment).
    \item[JSON:] answer in the JSON format.
    \item[Score:] answer as a number representing the positive class score.
    \item[Token-Word:] class probabilities calculated based on token probabilities, with answer as a single word.
    \item[Token-JSON:] class probabilities calculated based on token probabilities, with answer in the JSON format.
\end{description}

Two prompts were chosen based on the best average MCC scores from the previous experiment: B and F. They were appropriately modified for each of the output processing methods considered. The zero-shot setup was used and model output was limited to 256 tokens. For the output processing methods where tokens were actually sampled, deterministic sampling was used by setting the temperature parameter to zero.

The expected answers in JSON format were the following:
\begin{itemize}
    \item \texttt{\{"high\_quality": "false"\}} or \texttt{\{"high\_quality": "true"\}} for prompt~B,
    \item \texttt{\{"included": "false"\}} or \texttt{\{"included": "true"\}} for prompt~F. 
\end{itemize}
This difference is a consequence of the question asked in the prompts: prompt B asks whether the sample is a high-quality abstract on the topic, while prompt F asks whether the article this abstract describes should be included in a review on the topic.

These answers were searched for in the model output in the same way as in the previous experiment. The braces were ignored in order to recognize more model answers.

In the configuration with model output as a number representing the positive class probability, the model was prompted to give an integer number from 0 to 5, where a higher number meant that the given abstract was more likely to be of a high-quality article on the given topic. The result was then normalized to the $[0,1]$ range and treated as the probability of the positive class.

In the variants with calculation based on first token probabilities, only one token distribution was generated. Of this distribution, tokens representing text that could be a prefix of an expected answer (case insensitive) were selected. The probabilities of the tokens corresponding to each answer were summed and normalized so that they sum to 1. In the configuration where JSON format was used as well, the beginning of the answer (\texttt{\{"high\_quality":} for prompt B or \texttt{\{"included":} for prompt F) was given as part of the prompt in the assistant message and the expected answers were just \texttt{true} and \texttt{false}. This made it possible to identify the predicted class or estimate class probabilities based on the first generated output token.

Table~\ref{tab:dcr.output.auprc} contains the AUPRC scores for all compared output processing methods. The results in this table are macro-averaged over both prompts used. Additionally, Appendix~\ref{app:mcc} contains Table~\ref{tab:dcr.output.mcc} with the MCC scores averaged over both prompts, Appendix~\ref{app:models} contains Table~\ref{tab:dcr.output.avg} with scores for each model, for each considered output processing method and prompt averaged over all datasets, and Appendix~\ref{app:prc} contains Figure~\ref{fig:output:pr} with the precision-recall curves for each dataset, for each variant that outputs class probabilities. 

Processing model output by calculating class probabilities based on first token probabilities has proven to produce the best class probabilities, as shown by the AUPRC score being significantly higher for this output method, than for other tested methods.

Prompting the model to estimate positive class probability as a number in most cases achieved better AUPRCs than Word and JSON output processing methods, which means that the resulting probabilities are predictively useful.

\begin{table}[!htb]
\caption{Macro-averaged AUPRC scores for the Output experiment.}\label{tab:dcr.output.auprc}
\resizebox{\linewidth}{!}{%
\centering
\begin{tabular}{|c|c|c|c|c|c|}
\hline
Output & Word & JSON & Score & \textbf{Token-Word} & Token-JSON \\
\hline\hline
ACEInhibitors & $0.088$ & $0.088$ & $0.107$ & $\mathbf{0.173}$ & $0.153$ \\
\hline
ADHD & $0.352$ & $0.371$ & $0.386$ & $\mathbf{0.579}$ & $0.531$ \\
\hline
Antihistamines & $0.359$ & $0.367$ & $0.386$ & $\mathbf{0.496}$ & $0.477$ \\
\hline
AtypicalAntipsychotics & $0.349$ & $0.361$ & $0.389$ & $\mathbf{0.499}$ & $0.471$ \\
\hline
BetaBlockers & $0.171$ & $0.171$ & $0.195$ & $0.253$ & $\mathbf{0.255}$ \\
\hline
CalciumChannelBlockers & $0.256$ & $0.256$ & $0.274$ & $0.287$ & $\mathbf{0.291}$ \\
\hline
Estrogens & $0.257$ & $0.267$ & $0.308$ & $\mathbf{0.495}$ & $0.483$ \\
\hline
NSAIDS & $0.246$ & $0.259$ & $0.292$ & $\mathbf{0.426}$ & $0.410$ \\
\hline
Opiods & $0.029$ & $0.030$ & $0.038$ & $\mathbf{0.121}$ & $0.096$ \\
\hline
OralHypoglycemics & $0.305$ & $0.312$ & $0.328$ & $0.429$ & $\mathbf{0.435}$ \\
\hline
ProtonPumpInhibitors & $0.209$ & $0.220$ & $0.237$ & $\mathbf{0.338}$ & $0.321$ \\
\hline
SkeletalMuscleRelaxants & $0.060$ & $0.054$ & $0.073$ & $\mathbf{0.151}$ & $0.136$ \\
\hline
Statins & $0.070$ & $0.072$ & $0.091$ & $\mathbf{0.150}$ & $0.138$ \\
\hline
Triptans & $0.416$ & $0.442$ & $0.467$ & $\mathbf{0.666}$ & $0.625$ \\
\hline
UrinaryIncontinence & $0.257$ & $0.264$ & $0.282$ & $0.370$ & $\mathbf{0.375}$ \\
\hline
\hline
\textbf{Average} & $0.228$ & $0.236$ & $0.257$ & $\mathbf{0.362}$ & $0.346$ \\
\hline
\end{tabular}
}
\end{table}

\subsection{Few-Shot-Count Experiment}

In this experiment the effects of changing the number of examples used for few-shot prompts are investigated, with the zero-shot case also included for comparison. The training subset of each cross-validation fold is available for example selection, with the investigation of the effects of more limited example availability postponed for future work. Only small numbers of examples, between 1 and 4, are used because of the length of samples in the datasets, which causes long inference times and with higher number of examples would cause exceeding the context limit of one of the LLMs used (Gemma 2 9B Instruct). 

Examples are selected using the L2 similarity in the embedding space of the all-MiniLM-L6-v2 sentence transformers model. It is a fine-tuned model based on MiniLM~\cite{wang2020minilmdeepselfattentiondistillation} and it is recommended by the authors of the library as lightweight but providing good quality. The embedding vectors provided by this model are normalized to unit length, so the L2 and cosine similarity measures are equivalent.

Based on the best average AUPRC score achieved in the previous experiment, prompt B with Token-Word output processing method was selected for this experiment. The examples were provided to the model as repeated user and assistant messages, where the user message contained the article abstract and the assistant message contained the correct answer. This option was chosen over others like appending all examples to the system message in order to make the format of the actual output (contained in an assistant message) completely analogous to the examples and thus easier for the LLM to follow. The final user message contained the classified sample. The order of examples is determined by the similarity score -- the examples most similar to the classified sample are earlier in the prompt.

AUPRC scores for the experiment, for each dataset, are presented in Table~\ref{tab:dcr.n_shot.auprc}. Additionally, Table~\ref{tab:dcr.n_shot.mcc} in Appendix~\ref{app:mcc} contains the MCC scores, Table~\ref{tab:dcr.n_shot.avg} in Appendix~\ref{app:models} contains scores for each model macro-averaged over all datasets, and Figure~\ref{fig:n_shot:pr} in Appendix~\ref{app:prc} presents the precision-recall curves for each dataset, for each examples count.

The results for AUPRC score show that classification quality improves greatly when adding the first example. However, the improvements diminish with an increasing number of examples; the fourth example did not improve AUPRC at all.

Aside from classification quality, an important practical consideration is also the increase in computational power required with larger numbers of examples. With the rate of quality improvement decreasing for larger example counts, it might be better not to set the number of examples to the highest value the context limit permits.

\begin{table}[!htb]
\caption{Macro-averaged AUPRC scores for the Few-Shot-Count experiment.}\label{tab:dcr.n_shot.auprc}
\centering
\begin{tabular}{|c|c|c|c|c|c|}
\hline
Examples & 0 & 1 & 2 & \textbf{3} & 4 \\
\hline\hline
ACEInhibitors & $0.172$ & $0.210$ & $0.241$ & $\mathbf{0.264}$ & $0.232$ \\
\hline
ADHD & $0.551$ & $0.585$ & $0.580$ & $0.594$ & $\mathbf{0.601}$ \\
\hline
Antihistamines & $0.459$ & $0.487$ & $\mathbf{0.503}$ & $0.490$ & $0.463$ \\
\hline
AtypicalAntipsychotics & $0.520$ & $0.602$ & $0.622$ & $\mathbf{0.633}$ & $0.623$ \\
\hline
BetaBlockers & $0.246$ & $0.282$ & $0.311$ & $0.320$ & $\mathbf{0.330}$ \\
\hline
CalciumChannelBlockers & $0.297$ & $0.425$ & $0.478$ & $0.491$ & $\mathbf{0.510}$ \\
\hline
Estrogens & $0.481$ & $0.637$ & $0.625$ & $\mathbf{0.644}$ & $0.631$ \\
\hline
NSAIDS & $0.445$ & $0.517$ & $0.585$ & $0.617$ & $\mathbf{0.627}$ \\
\hline
Opiods & $0.079$ & $0.124$ & $0.157$ & $0.183$ & $\mathbf{0.186}$ \\
\hline
OralHypoglycemics & $0.437$ & $0.511$ & $0.552$ & $\mathbf{0.591}$ & $0.576$ \\
\hline
ProtonPumpInhibitors & $0.340$ & $0.437$ & $\mathbf{0.460}$ & $0.451$ & $0.452$ \\
\hline
SkeletalMuscleRelaxants & $0.166$ & $0.269$ & $0.251$ & $0.297$ & $\mathbf{0.343}$ \\
\hline
Statins & $0.163$ & $0.189$ & $0.211$ & $0.217$ & $\mathbf{0.234}$ \\
\hline
Triptans & $0.676$ & $0.685$ & $0.720$ & $\mathbf{0.725}$ & $0.710$ \\
\hline
UrinaryIncontinence & $0.417$ & $0.460$ & $0.488$ & $\mathbf{0.502}$ & $0.478$ \\
\hline
\hline
\textbf{Average} & $0.363$ & $0.428$ & $0.452$ & $\mathbf{0.468}$ & $0.466$ \\
\hline
\end{tabular}
\end{table}

\subsection{Few-Shot-Selection Experiment}

In this experiment, the following methods of example selection for few-shot prompts are compared:
\begin{description}
    \item[random:] random selection,
    \item[minilm:] similarity-based selection using the all-MiniLM-L6-v2 sentence transformers model,
    \item[distroberta:] similarity-based selection using the all-distilroberta-v1 sentence transformers model,
    \item[mpnet:] similarity-based selection using the all-mpnet-base-v2  sentence transformers model,
    \item[mpnet-bal:] similarity-based selection using the all-mpnet-base-v2  sentence transformers model with class rebalancing (this model was chosen based on the results of preliminary experiments).
\end{description}

Aside from the already mentioned all-MiniLM-L6-v2, the other embedding models are all-distilroberta-v1, a fine-tuned model based on DistilRoBERTa~\cite{Sanh2019DistilBERTAD} selected because of its longer supported context, and all-mpnet-base-v2, a fine-tuned model based on MPNet~\cite{song2020mpnetmaskedpermutedpretraining} selected as providing the best results among the Sentence Transformers models according to the authors of the library. The L2 similarity in the embedding space is used.

Three examples were used as this variant achieved the best average AUPRC score in the previous experiment. In the configuration with class rebalancing, there are always two negative class examples and one positive class example, as all the datasets are unbalanced in favor of the negative class. The order of examples is the same as in the previous experiment, except for the configuration with class balancing: there, the negative class examples are always first in the prompt.

The AUPRC scores for the experiment, for each dataset, are presented in Table~\ref{tab:dcr.select.auprc}. Additionally, Table~\ref{tab:dcr.select.mcc} in Appendix~\ref{app:mcc} contains the MCC scores for each dataset, Table~\ref{tab:dcr.select.avg} in Appendix~\ref{app:models} contains scores for each model macro-averaged over all datasets, and Figure~\ref{fig:select:pr} in Appendix~\ref{app:prc} presents the precision-recall curves for each dataset, for each example selection method.

The differences in quality between the three considered embedding models are quite small. However, all of them achieve considerably better results than random selection. Enforcing a class balance among the examples turned out not to be an effective strategy -- this configuration achieved worse results than all three variants with regular selection based on semantic similarity.

\begin{table}[!htb]
\caption{Macro-averaged AUPRC scores for the Few-Shot-Selection experiment.}\label{tab:dcr.select.auprc}
\resizebox{\linewidth}{!}{%
\centering
\begin{tabular}{|c|c|c|c|c|c|}
\hline
Method & random & minilm & \textbf{distroberta} & mpnet & mpnet-bal \\
\hline\hline
ACEInhibitors & $0.176$ & $0.265$ & $\mathbf{0.305}$ & $0.267$ & $0.155$ \\
\hline
ADHD & $0.584$ & $0.595$ & $0.647$ & $\mathbf{0.664}$ & $0.536$ \\
\hline
Antihistamines & $0.473$ & $0.490$ & $\mathbf{0.517}$ & $0.497$ & $0.434$ \\
\hline
AtypicalAntipsychotics & $0.500$ & $\mathbf{0.633}$ & $0.616$ & $0.625$ & $0.542$ \\
\hline
BetaBlockers & $0.247$ & $0.319$ & $0.327$ & $\mathbf{0.344}$ & $0.275$ \\
\hline
CalciumChannelBlockers & $0.295$ & $\mathbf{0.492}$ & $0.469$ & $0.479$ & $0.350$ \\
\hline
Estrogens & $0.402$ & $\mathbf{0.645}$ & $0.625$ & $0.626$ & $0.465$ \\
\hline
NSAIDS & $0.433$ & $\mathbf{0.617}$ & $0.608$ & $0.615$ & $0.511$ \\
\hline
Opiods & $0.100$ & $0.183$ & $\mathbf{0.260}$ & $0.180$ & $0.125$ \\
\hline
OralHypoglycemics & $0.457$ & $\mathbf{0.590}$ & $0.581$ & $0.570$ & $0.498$ \\
\hline
ProtonPumpInhibitors & $0.308$ & $0.451$ & $0.443$ & $\mathbf{0.460}$ & $0.362$ \\
\hline
SkeletalMuscleRelaxants & $0.161$ & $\mathbf{0.297}$ & $0.269$ & $0.253$ & $0.221$ \\
\hline
Statins & $0.144$ & $0.217$ & $0.235$ & $\mathbf{0.239}$ & $0.172$ \\
\hline
Triptans & $0.665$ & $0.726$ & $\mathbf{0.747}$ & $0.736$ & $0.668$ \\
\hline
UrinaryIncontinence & $0.405$ & $\mathbf{0.503}$ & $0.494$ & $0.495$ & $0.444$ \\
\hline
\hline
\textbf{Average} & $0.357$ & $0.468$ & $\mathbf{0.476}$ & $0.470$ & $0.384$ \\
\hline
\end{tabular}
}
\end{table}

In addition to that, Table~\ref{tab:dcr.select.exused} presents how many training samples were actually selected for the classification of any sample from the test set. Usually about 80\% of the training set was used, with this value being slightly lower for the variant with class balancing. This means that if examples were selected from a smaller random subset of the training set, classification quality would likely deteriorate. A potential avenue for further research is the evaluation of how classification performance would change in the case of a smaller training set.

\begin{table}[!htb]
\caption{Percent of training set selected as examples during the Few-Shot-Selection experiment.}\label{tab:dcr.select.exused}
\resizebox{\linewidth}{!}{%
\centering
\begin{tabular}{|c|c|c|c|c|c|}
\hline
Method & random & minilm & distroberta & mpnet & mpnet-bal \\
\hline\hline
ACEInhibitors & $95.7$ & $81.3$ & $81.3$ & $83.5$ & $76.8$ \\
\hline
ADHD & $95.4$ & $77.3$ & $82.1$ & $81.9$ & $76.6$ \\
\hline
Antihistamines & $93.4$ & $80.1$ & $80.8$ & $83.3$ & $84.7$ \\
\hline
AtypicalAntipsychotics & $95.2$ & $79.6$ & $79.4$ & $81.2$ & $80.6$ \\
\hline
BetaBlockers & $94.9$ & $80.7$ & $80.4$ & $82.7$ & $79.3$ \\
\hline
CalciumChannelBlockers & $94.5$ & $81.1$ & $82.1$ & $82.0$ & $80.1$ \\
\hline
Estrogens & $94.0$ & $79.7$ & $82.2$ & $84.2$ & $82.8$ \\
\hline
NSAIDS & $94.4$ & $77.9$ & $80.4$ & $81.3$ & $81.0$ \\
\hline
Opiods & $95.0$ & $80.5$ & $79.6$ & $82.2$ & $74.5$ \\
\hline
OralHypoglycemics & $94.7$ & $77.5$ & $78.9$ & $81.3$ & $81.7$ \\
\hline
ProtonPumpInhibitors & $95.0$ & $77.6$ & $78.5$ & $80.8$ & $78.9$ \\
\hline
SkeletalMuscleRelaxants & $95.0$ & $84.3$ & $83.3$ & $85.4$ & $76.5$ \\
\hline
Statins & $95.2$ & $80.9$ & $81.7$ & $83.3$ & $76.9$ \\
\hline
Triptans & $95.1$ & $80.0$ & $81.6$ & $81.8$ & $77.9$ \\
\hline
UrinaryIncontinence & $93.0$ & $83.8$ & $81.3$ & $84.2$ & $81.7$ \\
\hline
\end{tabular}
}
\end{table}

\subsection{Advanced-Techniques Experiment}

In this experiment some more advanced techniques of using LLM for text classification are investigated. These are:
\begin{itemize}
    \item chain of thought with single (\textbf{cot1}) or double (\textbf{cot2}) model requests,
    \item tree of thoughts (\textbf{tot}),
    \item chunking, with 0 (\textbf{chu}) and 3 (\textbf{chu-fs}) examples.
\end{itemize}

\subsubsection{Chain of Thought}

In this study, the chain of thought method is used in two ways: 
\begin{itemize}
    \item with a single LLM request,
    \item with two LLM requests
\end{itemize}
for each sample being classified.

In the variant with one request to the model, the model is prompted to first generate a reasoning and then generate the answer to the question, in one prompt. In order to easily distinguish the final answer from the reasoning, the model is expected to give the final answer after the reasoning in JSON format. This corresponds to the JSON output processing method.

In the variant with two requests to the model, first, the model is prompted to generate a JSON list of relevant phrases extracted from the original text that might point to it belonging to one of the classes. It is supposed to be a single list of phrases relevant to either of the classes. The phrases do not need to be exactly the same as in the original text, though the model was not prompted to modify them. This JSON list was detected using a regular expression and included in the prompt for the second call of the model alongside the original classified sample. The rest of the first request's output was discarded. There was some superfluous text discarded in over 70\% of model outputs. It usually contained some conclusions resulting from the extracted phrases, though there were cases where this resulted from misinterpretation of the output format and the model generating, for example, more than one JSON list. The objective of the second call of the model was then to actually classify the sample, using the JSON output processing method. Neither of the models was prompted to generate any additional reasoning.

All prompts in both variants are structured as follows: expected model role and information about task in system message, question about task and article abstract in user message, beginning of the expected answer in assistant message. For the single request variant, the assistant message contained the phrase ``Let's think step by step'', while in the two requests variant the assistant message contained the beginning of the expected JSON list or object. This prompt construction, aside from the assistant messages, is analogous to prompt F from previous experiments -- in the Prompts experiment it was the best prompt in average MCC score.

In both variants, the final answer is given in JSON format, with output processing method same as the JSON method in the Output experiment.

Only zero-shot variant is experimented with for these methods because of the difficulty of manually preparing examples of correct reasoning for DCR samples. Maximum number of output tokens was set to 256.

\subsubsection{Tree of Thoughts}

For the tree of thoughts technique the same steps were used as in chain of thought with two LLM requests: first step's objective is to generate a list of relevant phrases extracted from the text while the second step outputs the class of the sample. In order to achieve some variance in model outputs, models in both steps were prompted to generate a reasoning before the final answer.

Unlike for the previous experiments, here, in order to generate different answers during each call of the model at the same step, deterministic sampling was not used. Sampling parameters were set as follows:
\begin{description}
\item[temperature:] parameter controlling softmax token selection -- set to $0.8$,
\item[top\_k:] parameter restricting the number of the most probable tokens considered -- set to $40$ (default for llama-cpp-python),
\item[top\_p:] parameter restricting the cumulative probability of the most probable tokens to consider -- set to $0.95$ (default for llama-cpp-python),
\item[min\_p:] parameter restricting the minimum probability of tokens to consider relative to the most probable token -- $0.05$ (default for llama-cpp-python).
\end{description}

The answers for each step were scored in the range from 0 to 5. The threshold for accepting an answer was set to 3. The maximum number of answers generated for a step before backtracking to the previous step was set to 3. Maximum number of output tokens was set to 512 because of the long, but correct, reasonings that were noticed in the model's outputs.

Only the zero-shot variant was experimented with for these methods because of the difficulty of manually preparing examples of correct reasoning for DCR samples and the large amount of computing power this method requires.

All prompts for tree of thoughts, including prompts for generation and evaluation, were structured analogously to the prompts used for chain of thought. Evaluation prompts did not contain an explicit question about task, instead there was a command to rate the generated answer. The evaluation model was supposed to rate the quality of the previously generated answer on a scale from 0 to 5. Additionally, the JSON format was used to obtain the answer, as it was done for the generation itself.

\subsubsection{Chunking}

Samples were split into chunks with maximum size of 800 characters. Splitting was done only at word boundaries. Whitespace was removed from both the beginning and the end of the chunks.

In the few-shot variant, the examples were split as well, in order to make them analogous to the actual classified samples. Each chunk obtained from an example was considered to have the same class as the original example. Example selection using semantic similarity was done on a set of chunks obtained by splitting all the training samples.

The prompts were modified in order to inform the model that it is supposed to classify a fragment of an abstract, not the entire abstract. Aside from this change, the prompt was structured analogously to the prompt from the Few-Shot-Count experiment.

Model outputs were constrained to the expected answers only, to avoid the issue of unrecognized answers. This was done by setting the probability of tokens that did not fit any of the expected answers to zero. Model outputs obtained from chunks were aggregated via averaging and the average was treated as the probability of the sample belonging to the positive class.

\subsubsection{Results}

To compare advanced techniques with base LLM classification, the best few-shot variant from the previous experiments was used (prompt B, Token-Word output processing method, 3-shot, examples selection using L2 similarity in the embedding space of all-distilroberta-v1 model), labeled \textbf{base-fs}, as well as the best zero-shot variant (prompt B, Token-Word output processing method), labeled \textbf{base-zs}.

Tables~\ref{tab:dcr.advanced.auprc} and \ref{tab:dcr.advanced.mcc} contain respectively the AUPRC scores and MCC scores for each dataset. Additionally, Table~\ref{tab:dcr.advanced.avg} in Appendix~\ref{app:models} contains scores for each model averaged over all datasets. Precision-recall curves are not presented because among the considered advanced methods only chunking returns any kind of class probabilities, and even these probabilities are limited to multiples of $\frac{1}{n}$, where $n$ is the number of chunks the classified sample was split into. Overall, all of the advanced methods experimented with here failed to achieve better AUPRC scores than classification using output processing via calculating class probabilities based on token probabilities from Few-Shot-Selection experiment. However, the results of the advanced methods are better in the case of MCC score. This might result from the fact that the variant from Few-Shot-Selection experiment was selected because of its best AUPRC score, not MCC score. Still, considering the AUPRC score as more important because of its consideration of all possible probability thresholds, the advanced methods failed to compensate their larger computing power cost.

\begin{table}[!htb]
\caption{Macro-averaged AUPRC scores for the Advanced-Techniques experiment.}\label{tab:dcr.advanced.auprc}
\resizebox{\linewidth}{!}{%
\centering
\begin{tabular}{|c|c|c|c|c|c|c|c|}
\hline
Variant & base-zs & \textbf{base-fs} & chu & chu-fs & cot1 & cot2 & tot \\
\hline\hline
ACEInhibitors & $0.172$ & $\mathbf{0.305}$ & $0.099$ & $0.093$ & $0.103$ & $0.092$ & $0.096$ \\
\hline
ADHD & $0.551$ & $\mathbf{0.647}$ & $0.398$ & $0.353$ & $0.426$ & $0.377$ & $0.259$ \\
\hline
Antihistamines & $0.459$ & $\mathbf{0.517}$ & $0.382$ & $0.376$ & $0.385$ & $0.363$ & $0.353$ \\
\hline
AtypicalAntipsychotics & $0.520$ & $\mathbf{0.616}$ & $0.370$ & $0.367$ & $0.373$ & $0.375$ & $0.373$ \\
\hline
BetaBlockers & $0.246$ & $\mathbf{0.327}$ & $0.193$ & $0.184$ & $0.182$ & $0.175$ & $0.174$ \\
\hline
CalciumChannelBlockers & $0.297$ & $\mathbf{0.469}$ & $0.258$ & $0.256$ & $0.258$ & $0.260$ & $0.258$ \\
\hline
Estrogens & $0.481$ & $\mathbf{0.625}$ & $0.274$ & $0.285$ & $0.302$ & $0.262$ & $0.269$ \\
\hline
NSAIDS & $0.445$ & $\mathbf{0.608}$ & $0.266$ & $0.271$ & $0.277$ & $0.275$ & $0.291$ \\
\hline
Opiods & $0.079$ & $\mathbf{0.260}$ & $0.036$ & $0.030$ & $0.036$ & $0.029$ & $0.028$ \\
\hline
OralHypoglycemics & $0.437$ & $\mathbf{0.581}$ & $0.322$ & $0.328$ & $0.332$ & $0.314$ & $0.320$ \\
\hline
ProtonPumpInhibitors & $0.340$ & $\mathbf{0.443}$ & $0.238$ & $0.217$ & $0.236$ & $0.218$ & $0.224$ \\
\hline
SkeletalMuscleRelaxants & $0.166$ & $\mathbf{0.269}$ & $0.074$ & $0.070$ & $0.058$ & $0.048$ & $0.055$ \\
\hline
Statins & $0.163$ & $\mathbf{0.235}$ & $0.079$ & $0.078$ & $0.082$ & $0.076$ & $0.077$ \\
\hline
Triptans & $0.676$ & $\mathbf{0.747}$ & $0.454$ & $0.441$ & $0.466$ & $0.457$ & $0.451$ \\
\hline
UrinaryIncontinence & $0.417$ & $\mathbf{0.494}$ & $0.253$ & $0.273$ & $0.259$ & $0.287$ & $0.294$ \\
\hline
\hline
\textbf{Average} & $0.363$ & $\mathbf{0.476}$ & $0.246$ & $0.241$ & $0.252$ & $0.241$ & $0.235$ \\
\hline
\end{tabular}
}
\end{table}

\begin{table}[!htb]
\caption{Macro-averaged MCC scores for the Advanced-Techniques experiment.}\label{tab:dcr.advanced.mcc}
\resizebox{\linewidth}{!}{%
\centering
\begin{tabular}{|c|c|c|c|c|c|c|c|}
\hline
Variant & base-zs & base-fs & chu & chu-fs & \textbf{cot1} & cot2 & tot \\
\hline\hline
ACEInhibitors & $0.099$ & $0.104$ & $0.106$ & $0.097$ & $\mathbf{0.136}$ & $0.102$ & $0.107$ \\
\hline
ADHD & $0.470$ & $0.472$ & $0.481$ & $0.448$ & $\mathbf{0.540}$ & $0.520$ & $0.349$ \\
\hline
Antihistamines & $0.214$ & $0.190$ & $\mathbf{0.282}$ & $0.235$ & $0.259$ & $0.207$ & $0.141$ \\
\hline
AtypicalAntipsychotics & $0.153$ & $0.144$ & $0.181$ & $0.174$ & $0.157$ & $\mathbf{0.195}$ & $0.163$ \\
\hline
BetaBlockers & $0.129$ & $0.130$ & $0.145$ & $0.130$ & $\mathbf{0.146}$ & $0.132$ & $0.118$ \\
\hline
CalciumChannelBlockers & $\mathbf{0.134}$ & $0.121$ & $0.117$ & $0.124$ & $0.095$ & $0.121$ & $0.103$ \\
\hline
Estrogens & $0.194$ & $0.154$ & $0.198$ & $0.217$ & $\mathbf{0.246}$ & $0.180$ & $0.163$ \\
\hline
NSAIDS & $0.138$ & $0.112$ & $0.130$ & $0.137$ & $0.159$ & $\mathbf{0.199}$ & $0.183$ \\
\hline
Opiods & $0.062$ & $0.049$ & $0.078$ & $0.050$ & $\mathbf{0.091}$ & $0.047$ & $0.021$ \\
\hline
OralHypoglycemics & $0.136$ & $0.144$ & $0.112$ & $0.125$ & $\mathbf{0.166}$ & $0.124$ & $0.114$ \\
\hline
ProtonPumpInhibitors & $0.136$ & $0.121$ & $0.152$ & $0.089$ & $\mathbf{0.160}$ & $0.143$ & $0.144$ \\
\hline
SkeletalMuscleRelaxants & $\mathbf{0.183}$ & $0.181$ & $0.170$ & $0.170$ & $0.147$ & $0.130$ & $0.139$ \\
\hline
Statins & $0.110$ & $0.117$ & $0.129$ & $0.125$ & $\mathbf{0.136}$ & $0.121$ & $0.118$ \\
\hline
Triptans & $0.313$ & $0.274$ & $0.329$ & $0.295$ & $0.332$ & $\mathbf{0.357}$ & $0.305$ \\
\hline
UrinaryIncontinence & $0.098$ & $0.131$ & $0.081$ & $0.128$ & $0.056$ & $0.144$ & $\mathbf{0.160}$ \\
\hline
\hline
\textbf{Average} & $0.171$ & $0.163$ & $0.179$ & $0.170$ & $\mathbf{0.188}$ & $0.181$ & $0.155$ \\
\hline
\end{tabular}
}
\end{table}

\subsection{Baseline Experiment}

Two conventional classification algorithm were used as baseline for this study: random forest and multinomial na\"{i}ve Bayes. For both algorithms, the scikit-learn implementation was used. Text representation for these algorithms was obtained with the bag of words model, using the scikit-learn implementation (CountVectorizer). Stop word removal was performed using the English stop words list from nltk~\cite{bird2009nltk} library. In addition to that, words contained in less than 3 training samples were removed from the vocabulary.

In addition to that, two fine-tuned transformer models were selected as baselines: DeBERTa~\cite{he2021deberta} and SciDeBERTa-v2~\cite{medibiodeberta}. DeBERTa has been used as baseline in several prior studies on LLM text classification~\cite{sun23classification,wang23zeroshot,chae23classification,wang25adaptable}, while SciDeBERTa-v2, as a variant of DeBERTa-v2 pre-trained on scientific texts, can be expected to achieve better results on Drug Class Review datasets. Both models were downloaded and trained using HuggingFace libraries.

The evaluation for the algorithms described above was run using stratified 5-fold cross validation. The given results are macro-averaged from all runs. Na\"{i}ve Bayes was run using default hyperparameters from scikit-learn library. For random forest, results are presented for both default parameters (except the number of estimators, which was set to 1000) and for parameters tuned using grid search. The final evaluation of random forest was run 25 times for different random seeds (hyperparameter optimization was not repeated); the given results are macro-averaged.

Three hyperparameters of random forest were optimized using grid search with non-repeated 5-fold stratified cross validation based on highest value of macro-averaged AUPRC, separately for each loop of the main evaluation's cross validation. The optimized hyperparameters, their names in the scikit-learn implementation, and their values that were considered are the following:
\begin{itemize}
    \item maximum tree depth (max\_depth): 20, 50, unlimited;
    \item maximum number of features considered (max\_features): $\lfloor\sqrt{n}\rfloor$, $\lfloor\text{log}_2(n)\rfloor$, 0.2, 0.3, 0.4, 0.5 (where $n$ is the total number of features);
    \item class weights (class\_weight): balanced (inversely proportional to class frequencies in the training set), equal for all classes.
\end{itemize}

Transformer models have been fine-tuned with 5 epochs, $2\cdot 10^{-5}$ learning rate, $0.01$ weight decay and a batch size of 4, except for the Statins dataset for which batch size 2 was used because of several extremely long samples not fitting in GPU memory. The rest of the training configuration was left as default:
\begin{itemize}
    \item the classification head was a linear layer on top of the pooled model output,
    \item no model weights were frozen,
    \item AdamW optimizer was used with $\beta_1=0.9$ and $\beta_2=0.999$,
    \item a linear learning rate scheduler with no warmup steps was used,
    \item the maximum gradient norm was set to $1.0$,
    \item the dropout probability was set to $0.1$.
\end{itemize}

For comparison of LLMs with conventional algorithms, the best variant selected in the Few-Shot-Selection experiment was used (prompt B, Token-Word output processing method, 3-shot, examples selection using L2 similarity in the embedding space of all-distilroberta-v1 model).

In order to also explore the differences in text classification capabilities between openly available and commercial models, two additional models were included in this comparison: GPT-4.1 mini and GPT-4.1 nano. The OpenAI Responses API makes the probabilities of only the top 20 tokens available. For the Token-Word output processing method, the probabilities of all other tokens were assumed to be zero. The class probabilities were calculated based on this probability distribution of the first output token in the same way as for the other models.

To summarize, the following classifiers are compared:
\begin{description}
    \item[gemm:] the best variant chosen based on the Few-Shot-Selection experiment with the Gemma 2 9B Instruct model,
    \item[ll-8b:] the best variant chosen based on the Few-Shot-Selection experiment with the Llama 3.1 8B Instruct model,
    \item[ll-70b:] the best variant chosen based on the Few-Shot-Selection with the Llama 3.1 70B Instruct model,
    \item[mistr:] best variant chosen based on the Few-Shot-Selection experiment with the Mistral 7B Instruct v0.3 model,
    \item[gptmi:] the best variant chosen based on the Few-Shot-Selection experiment with the GPT-4.1 mini model (version: gpt-4\_1-mini-2025-04-14),
    \item[gptna:] the best variant chosen based on the Few-Shot-Selection experiment with the GPT-4.1 nano model (version: gpt-4\_1-nano-2025-04-14),
    \item[nb:] na\"{i}ve Bayes,
    \item[rf-d:] random forest with default hyperparameters,
    \item[rf-t:] random forest with tuned hyperparameters,
    \item[debe:] fine-tuned DeBERTa model,
    \item[scid:] fine-tuned SciDeBERTa-v2 model.
\end{description}

Table~\ref{tab:dcr.final.auprc} contains the AUPRC scores for each dataset. Additionally, Table~\ref{tab:dcr.final.mcc} in Appendix~\ref{app:mcc} contains the MCC scores for each dataset, Table~\ref{tab:dcr.final.avg} in Appendix~\ref{app:models} contains all performance scores averaged over all datasets, and Figure~\ref{fig:final:pr} in Appendix~\ref{app:prc} presents the precision-recall curves for each dataset, for each model type.

\begin{table}[!htb]
\caption{Macro-averaged AUPRC scores for the Baseline experiment.}\label{tab:dcr.final.auprc}
\resizebox{\linewidth}{!}{%
\centering
\begin{tabular}{|c|c|c|c|c|c|c|c|c|c|c|c|}
\hline
Alg & gemm & ll-70b & ll-8b & mistr & gptmi & gptna & nb & rf-d & \textbf{rf-t} & debe & scid \\
\hline\hline
Ace & $0.323$ & $0.304$ & $0.341$ & $0.252$ & $0.113$ & $0.205$ & $0.314$ & $0.310$ & $0.387$ & $\mathbf{0.429}$ & $0.402$ \\
\hline
Adh & $\mathbf{0.693}$ & $0.676$ & $0.684$ & $0.534$ & $0.572$ & $0.594$ & $0.519$ & $0.681$ & $0.643$ & $0.549$ & $0.671$ \\
\hline
Ant & $0.485$ & $0.504$ & $0.572$ & $0.506$ & $0.402$ & $0.564$ & $0.548$ & $0.547$ & $\mathbf{0.640}$ & $0.529$ & $0.408$ \\
\hline
Aty & $0.620$ & $0.615$ & $0.624$ & $0.604$ & $0.497$ & $0.575$ & $0.667$ & $0.638$ & $0.674$ & $0.593$ & $\mathbf{0.678}$ \\
\hline
Bet & $0.292$ & $0.337$ & $0.349$ & $0.329$ & $0.222$ & $0.322$ & $0.400$ & $0.349$ & $0.416$ & $0.384$ & $\mathbf{0.454}$ \\
\hline
Cal & $0.441$ & $0.486$ & $0.491$ & $0.458$ & $0.314$ & $0.374$ & $0.569$ & $0.663$ & $\mathbf{0.686}$ & $0.510$ & $0.630$ \\
\hline
Est & $0.621$ & $\mathbf{0.656}$ & $0.578$ & $0.646$ & $0.401$ & $0.572$ & $0.618$ & $0.629$ & $0.652$ & $0.440$ & $0.575$ \\
\hline
Nsa & $0.601$ & $0.529$ & $0.652$ & $0.650$ & $0.368$ & $0.618$ & $0.703$ & $0.715$ & $\mathbf{0.759}$ & $0.701$ & $0.724$ \\
\hline
Opi & $0.236$ & $0.301$ & $0.277$ & $0.227$ & $0.054$ & $0.111$ & $0.095$ & $0.137$ & $0.255$ & $0.282$ & $\mathbf{0.349}$ \\
\hline
Ora & $0.572$ & $0.611$ & $0.567$ & $0.573$ & $0.447$ & $0.547$ & $0.606$ & $0.601$ & $0.603$ & $0.553$ & $\mathbf{0.618}$ \\
\hline
Pro & $0.437$ & $0.467$ & $0.441$ & $0.427$ & $0.310$ & $0.452$ & $0.479$ & $0.485$ & $\mathbf{0.540}$ & $0.423$ & $0.486$ \\
\hline
Ske & $0.266$ & $0.319$ & $0.335$ & $0.155$ & $0.188$ & $0.265$ & $0.303$ & $0.314$ & $\mathbf{0.343}$ & $0.111$ & $0.263$ \\
\hline
Sta & $0.234$ & $0.248$ & $0.254$ & $0.205$ & $0.163$ & $0.194$ & $0.315$ & $0.246$ & $\mathbf{0.342}$ & $0.247$ & $0.287$ \\
\hline
Tri & $0.752$ & $0.751$ & $0.756$ & $0.729$ & $0.519$ & $0.678$ & $0.723$ & $0.749$ & $\mathbf{0.764}$ & $0.742$ & $0.753$ \\
\hline
Uri & $0.460$ & $0.472$ & $0.497$ & $0.549$ & $0.451$ & $0.510$ & $0.621$ & $0.638$ & $0.641$ & $0.610$ & $\mathbf{0.670}$ \\
\hline
\hline
Avg & $0.469$ & $0.485$ & $0.495$ & $0.456$ & $0.335$ & $0.439$ & $0.499$ & $0.513$ & $\mathbf{0.556}$ & $0.474$ & $0.531$ \\
\hline
\end{tabular}
}
\end{table}

The baseline conventional algorithms turned out to be, on average, better than all of the considered LLMs. This is also the case for fine-tuned SciDeBERTa-v2 but not for DeBERTa which performed comparably to the best LLM setups. Both the transformer-based classifiers did not prove to provide substantial advantages over the conventional ones. The latter delivered superior classification quality for most datasets. However, for some datasets: ADHD, Estrogens, Opiods the AUPRC values achieved by LLMs were comparable or better. These datasets do not appear to be significantly different from the others with respect to statistics such as dataset size, class balance or average word length as presented in Table~\ref{tab:dcr.stats}.

Among the LLMs, the Llama 3.1 models achieved the best average AUPRC score. Surprisingly, the 8B variant performed slightly better than the much larger 70B variant, though Llama 3.1 70B achieved better results in all other performance scores.

The commercial LLMs did not achieve significantly different results from most of the openly available models used. A noticeable anomaly is the smaller and cheaper model GPT-4.1 nano having better AUPRC score than GPT-4.1 mini on all datasets.

\subsection{Discussion}

The main findings from the presented results are summarized below.
\begin{itemize}
\item While there is no clear relationship between prompt complexity and classification performance, it appears to be beneficial to include a direct question about the class of a given text sample the user message rather than just the text sample alone.
\item Processing output token probabilities makes it possible to obtain more reliable class probabilities that requesting the model to produce a numeric score, which permits controlling the precision-recall tradeoff by adjusting the decision threshold.
\item Providing examples considerably improves classification quality, with even one-shot prompts performing much better than zero-shot prompts, but there appears to be little or no benefit in using more than a few.
\item Similarity-based example selection for few-shot prompts is clearly better than random selection but the exact choice of the embedding model used to determine similarity is not essential. Rebalancing classes during example selection does not help.
\item There is no substantial improvement provided by two-step inference using more sophisticated chain or tree of thought prompting techniques, nor by chunking the text to be classified. 
\item While LLM-based text classifiers usually fail to outperform conventionally trained classification models, they come close, particularly when used in the few-shot setting.
\item There no clear and consistent performance differences among LLMs of different size and from different providers, with smaller open-source models not necessarily performing worse than larger commercial models in the text classification task. 
\end{itemize}

These findings directly lead to practical recommendations for the most useful LLM text classification setups. Whenever some labeled examples are available, it is preferable to select a few of them based on similarity to the text being classified and use them in the prompt, but not necessarily spend a lot of resources to process more than a handful for a single classification request. Whenever output token probabilities are available, it makes sense to use them to estimate class probabilities. There is no need to work out sophisticated prompting techniques but it pays off to include a direct question about the class in the user message part of the prompt, besides the task specification in the system message.

Unfortunately, no clear recommendations can be provided regarding prompt complexity or model selection. The relative performance of different prompts and models is likely to vary over different classification tasks. It is also possible that individually tuning prompts for particular models might lead to some noteworthy improvement. Exploring this possibility is postponed for future work. Similarly, it still remains to be verified how few-shot LLM-based text classifiers perform with a limited availability of labeled examples to choose from.

\section{Conclusion}

In this study, several setups of LLM-based medical article  classification were experimented with. Unlike most of existing literature, this work explores varied output processing methods, including obtaining the answer in JSON format, prompting the model to score the sample and calculating the positive class probability based on first token probabilities. Also unlike most existing work, in this article several example selection methods were compared, including three different embedding models. Aside from that, several prompts of varied complexity were considered, several examples counts were experimented with and a few more advanced techniques were explored: chain of thought, tree of thoughts, and sample chunking. While the set of LLMs experimented with is limited due to computational power and budget constraints, it includes four small and mid-size open source models, and in the final comparison experiment two closed source models are also used.

The obtained results show that output processing by calculating class probabilities based on token probabilities is promising despite its absence from existing literature. The superiority of the few-shot setup and of selecting examples based on semantic similarity have been confirmed. The more advanced techniques have proven ineffective despite their greater computing power requirements. The best few-shot configurations, while still not exceeding the classification quality achieved by  conventional classification algorithms for most datasets, are actually quite comparable, and even in the the zero-shot setting useful predictions are obtained. It remains to be seen whether newer or larger large language models, or different methods of their application, will turn out more useful in the difficult task of medical articles classification.

Possible future work could include a wider array of LLMs, including more closed-source ones, and a more diverse set of prompts, possibly individually tuned for particular models. More varied embedding models and embeddings based on LLMs themselves could be utilized. Other output processing methods could be researched, for example training an additional classifier on the internal embeddings of the LLM. With more computing power, larger examples counts could be experimented with. Other advanced methods of LLM utilization or including a few-shot variant of chain of thought and exploring more example selection methods, as well as fine-tuning LLMs to the specific classification task could be studied. Last but not least, it would be worthwhile to investigate the classification performance with few-shot prompts when the set of labeled text samples available for example selection is very limited. This would verify whether LLMs can be more competitive to conventional classification algorithms with scarce training data.

\section*{Acknowledgements}

We gratefully acknowledge Poland's high-performance infrastructure PLGrid ACK Cyfronet AGH for providing computer facilities and support within computational grant number PLG/2025/018275.

\newpage

\printbibliography

@inproceedings{sun23classification,
  author = {Sun, X. and Li, X. and Li, J. and Wu, F. and Guo, S. and Zhang, T. and Wang, G.},
  year = {2023},    
  title = {Text Classification via Large Language Models},
  editor = {Bouamor, H. and Pino, J. and Bali, K},
  booktitle = {Findings of the Association for Computational Linguistics: EMNLP 2023},
  publisher = {Association for Computational Linguistics},
  OPTurl = {https://aclanthology.org/2023.findings-emnlp.603/},
  OPTdoi = {10.18653/v1/2023.findings-emnlp.603},
  pages = {8990--9005}}

@misc{wang23zeroshot,
  author={Wang, Zhiqiang and Pang, Yiran and Lin, Yanbin},
  year = {2023},
  title={Large Language Models Are Zero-Shot Text Classifiers},
  howpublished = {arXiv:2312.01044},
  url={https://arxiv.org/abs/2312.01044}}

@article{chae23classification,
author = {Youngjin Chae and Thomas Davidson},
title ={Large Language Models for Text Classification: From Zero-Shot Learning to Instruction-Tuning},
journal = {Sociological Methods \& Research},
pages = {00491241251325243},
year = {2025},
OPTdoi = {10.1177/00491241251325243},
OPTurl = {https://doi.org/10.1177/00491241251325243},
}

@inproceedings{wang25adaptable,
  author = { Wang, Z. and Pang, Y. and Lin, Y. and Zhu, X.},
  year = {2024},
  title = {Adaptable and Reliable Text Classification using Large Language Models},
  booktitle = {Proceedings of the 2024 IEEE International Conference on Data Mining Workshops},
  series = {ICDMW-2024},
  pages = {67-74},
  publisher = {IEEE Computer Society},
  address = {Los Alamitos, CA, USA},
  OPTdoi = {10.1109/ICDMW65004.2024.00015},
  OPTurl = {https://doi.ieeecomputersociety.org/10.1109/ICDMW65004.2024.00015}}

@misc{edwards24textclassification,
      title = {Language Models for Text Classification: Is In-Context Learning Enough?}, 
      author = {Edwards, A. and Camacho-Collados, J.},
      year = {2024},
      howpublished = {arXiv:2403.17661},
      url={https://arxiv.org/abs/2403.17661}}

@article{guo24evaluating,
  author = {Guo, Y. and Ovadje, A. and Al-Garadi, M. A. and Sarker, A.},
  year = {2024},
  title = {Evaluating Large Language Models for Health-Related Text Classification Task with Public Social Media Data},
  journal = {Journal of the American Medical Informatics Association},
  volume = {31},
  number = {10},
  pages = {2181--2189},
  OPTurl = {https://doi.org/10.1093/jamia/ocae210}}

@misc{labrak24zeroshot,
  author = {Labrak, Y. and Rouvier, M. and Dufour, R.},
  year = {2024},
  title = {A Zero-Shot and Few-Shot Study of Instruction-Finetuned Large Language Models Applied to Clinical and Biomedical Tasks},
  howpublished = {arXiv:2307.12114},
  url={https://arxiv.org/abs/2307.12114}}

@article{rathje24gpt,
  author = {Rathje, S. and Mirea, D.-M.  and Sucholutsky, I.  and Marjieh, R.  and Robertson, C. E.  and Van Bavel, J. J.},
  year = {2024},
  title = {{GPT} is an Effective Tool for Multilingual Psychological Text Analysis},
  journal = {Proceedings of the National Academy of Sciences},
  volume = {121},
  number = {34},
  pages = {e2308950121},
  OPTdoi = {10.1073/pnas.2308950121},
  OPTurl = {https://www.pnas.org/doi/abs/10.1073/pnas.2308950121}}

@article{chen24evaluating,
  author = {Chen, S. and Li, Y. and Lu, S. and Van, H. and Aerts, H. J. W. L. and Savova, G. K. and Bitterman, D. S.},
  year = {2024},
  title = {Evaluating the {ChatGPT} Family of Models for Biomedical Reasoning and Classification},
  journal = {Journal of the American Medical Informatics Association},
  volume = {31},
  number = {4},
  pages = {940--948},
  OPTdoi = {10.1093/jamia/ocad256},
  OPTurl = {https://doi.org/10.1093/jamia/ocad256}}

@misc{mistral7b2023,
      title={Mistral {7B}}, 
      author={Albert Q. Jiang and others},
      FULLauthor={Albert Q. Jiang and Alexandre Sablayrolles and Arthur Mensch and Chris Bamford and Devendra Singh Chaplot and Diego de las Casas and Florian Bressand and Gianna Lengyel and Guillaume Lample and Lucile Saulnier and Lélio Renard Lavaud and Marie-Anne Lachaux and Pierre Stock and Teven Le Scao and Thibaut Lavril and Thomas Wang and Timothée Lacroix and William El Sayed},
      year={2023},
      howpublished={arXiv:2310.06825},
      url={https://arxiv.org/abs/2310.06825}, 
}

@misc{grattafiori2024llama3herdmodels,
      title={The {Llama 3} Herd of Models}, 
      author={Aaron Grattafiori and others},
      FULLauthor={Aaron Grattafiori and Abhimanyu Dubey and Abhinav Jauhri and Abhinav Pandey and Abhishek Kadian and Ahmad Al-Dahle and Aiesha Letman and Akhil Mathur and Alan Schelten and Alex Vaughan and Amy Yang and Angela Fan and Anirudh Goyal and Anthony Hartshorn and Aobo Yang and Archi Mitra and Archie Sravankumar and Artem Korenev and Arthur Hinsvark and Arun Rao and Aston Zhang and Aurelien Rodriguez and Austen Gregerson and Ava Spataru and Baptiste Roziere and Bethany Biron and Binh Tang and Bobbie Chern and Charlotte Caucheteux and Chaya Nayak and Chloe Bi and Chris Marra and Chris McConnell and Christian Keller and Christophe Touret and Chunyang Wu and Corinne Wong and Cristian Canton Ferrer and Cyrus Nikolaidis and Damien Allonsius and Daniel Song and Danielle Pintz and Danny Livshits and Danny Wyatt and David Esiobu and Dhruv Choudhary and Dhruv Mahajan and Diego Garcia-Olano and Diego Perino and Dieuwke Hupkes and Egor Lakomkin and Ehab AlBadawy and Elina Lobanova and Emily Dinan and Eric Michael Smith and Filip Radenovic and Francisco Guzmán and Frank Zhang and Gabriel Synnaeve and Gabrielle Lee and Georgia Lewis Anderson and Govind Thattai and Graeme Nail and Gregoire Mialon and Guan Pang and Guillem Cucurell and Hailey Nguyen and Hannah Korevaar and Hu Xu and Hugo Touvron and Iliyan Zarov and Imanol Arrieta Ibarra and Isabel Kloumann and Ishan Misra and Ivan Evtimov and Jack Zhang and Jade Copet and Jaewon Lee and Jan Geffert and Jana Vranes and Jason Park and Jay Mahadeokar and Jeet Shah and Jelmer van der Linde and Jennifer Billock and Jenny Hong and Jenya Lee and Jeremy Fu and Jianfeng Chi and Jianyu Huang and Jiawen Liu and Jie Wang and Jiecao Yu and Joanna Bitton and Joe Spisak and Jongsoo Park and Joseph Rocca and Joshua Johnstun and Joshua Saxe and Junteng Jia and Kalyan Vasuden Alwala and Karthik Prasad and Kartikeya Upasani and Kate Plawiak and Ke Li and Kenneth Heafield and Kevin Stone and Khalid El-Arini and Krithika Iyer and Kshitiz Malik and Kuenley Chiu and Kunal Bhalla and Kushal Lakhotia and Lauren Rantala-Yeary and Laurens van der Maaten and Lawrence Chen and Liang Tan and Liz Jenkins and Louis Martin and Lovish Madaan and Lubo Malo and Lukas Blecher and Lukas Landzaat and Luke de Oliveira and Madeline Muzzi and Mahesh Pasupuleti and Mannat Singh and Manohar Paluri and Marcin Kardas and Maria Tsimpoukelli and Mathew Oldham and Mathieu Rita and Maya Pavlova and Melanie Kambadur and Mike Lewis and Min Si and Mitesh Kumar Singh and Mona Hassan and Naman Goyal and Narjes Torabi and Nikolay Bashlykov and Nikolay Bogoychev and Niladri Chatterji and Ning Zhang and Olivier Duchenne and Onur Çelebi and Patrick Alrassy and Pengchuan Zhang and Pengwei Li and Petar Vasic and Peter Weng and Prajjwal Bhargava and Pratik Dubal and Praveen Krishnan and Punit Singh Koura and Puxin Xu and Qing He and Qingxiao Dong and Ragavan Srinivasan and Raj Ganapathy and Ramon Calderer and Ricardo Silveira Cabral and Robert Stojnic and Roberta Raileanu and Rohan Maheswari and Rohit Girdhar and Rohit Patel and Romain Sauvestre and Ronnie Polidoro and Roshan Sumbaly and Ross Taylor and Ruan Silva and Rui Hou and Rui Wang and Saghar Hosseini and Sahana Chennabasappa and Sanjay Singh and Sean Bell and Seohyun Sonia Kim and Sergey Edunov and Shaoliang Nie and Sharan Narang and Sharath Raparthy and Sheng Shen and Shengye Wan and Shruti Bhosale and Shun Zhang and Simon Vandenhende and Soumya Batra and Spencer Whitman and Sten Sootla and Stephane Collot and Suchin Gururangan and Sydney Borodinsky and Tamar Herman and Tara Fowler and Tarek Sheasha and Thomas Georgiou and Thomas Scialom and Tobias Speckbacher and Todor Mihaylov and Tong Xiao and Ujjwal Karn and Vedanuj Goswami and Vibhor Gupta and Vignesh Ramanathan and Viktor Kerkez and Vincent Gonguet and Virginie Do and Vish Vogeti and Vítor Albiero and Vladan Petrovic and Weiwei Chu and Wenhan Xiong and Wenyin Fu and Whitney Meers and Xavier Martinet and Xiaodong Wang and Xiaofang Wang and Xiaoqing Ellen Tan and Xide Xia and Xinfeng Xie and Xuchao Jia and Xuewei Wang and Yaelle Goldschlag and Yashesh Gaur and Yasmine Babaei and Yi Wen and Yiwen Song and Yuchen Zhang and Yue Li and Yuning Mao and Zacharie Delpierre Coudert and Zheng Yan and Zhengxing Chen and Zoe Papakipos and Aaditya Singh and Aayushi Srivastava and Abha Jain and Adam Kelsey and Adam Shajnfeld and Adithya Gangidi and Adolfo Victoria and Ahuva Goldstand and Ajay Menon and Ajay Sharma and Alex Boesenberg and Alexei Baevski and Allie Feinstein and Amanda Kallet and Amit Sangani and Amos Teo and Anam Yunus and Andrei Lupu and Andres Alvarado and Andrew Caples and Andrew Gu and Andrew Ho and Andrew Poulton and Andrew Ryan and Ankit Ramchandani and Annie Dong and Annie Franco and Anuj Goyal and Aparajita Saraf and Arkabandhu Chowdhury and Ashley Gabriel and Ashwin Bharambe and Assaf Eisenman and Azadeh Yazdan and Beau James and Ben Maurer and Benjamin Leonhardi and Bernie Huang and Beth Loyd and Beto De Paola and Bhargavi Paranjape and Bing Liu and Bo Wu and Boyu Ni and Braden Hancock and Bram Wasti and Brandon Spence and Brani Stojkovic and Brian Gamido and Britt Montalvo and Carl Parker and Carly Burton and Catalina Mejia and Ce Liu and Changhan Wang and Changkyu Kim and Chao Zhou and Chester Hu and Ching-Hsiang Chu and Chris Cai and Chris Tindal and Christoph Feichtenhofer and Cynthia Gao and Damon Civin and Dana Beaty and Daniel Kreymer and Daniel Li and David Adkins and David Xu and Davide Testuggine and Delia David and Devi Parikh and Diana Liskovich and Didem Foss and Dingkang Wang and Duc Le and Dustin Holland and Edward Dowling and Eissa Jamil and Elaine Montgomery and Eleonora Presani and Emily Hahn and Emily Wood and Eric-Tuan Le and Erik Brinkman and Esteban Arcaute and Evan Dunbar and Evan Smothers and Fei Sun and Felix Kreuk and Feng Tian and Filippos Kokkinos and Firat Ozgenel and Francesco Caggioni and Frank Kanayet and Frank Seide and Gabriela Medina Florez and Gabriella Schwarz and Gada Badeer and Georgia Swee and Gil Halpern and Grant Herman and Grigory Sizov and Guangyi and Zhang and Guna Lakshminarayanan and Hakan Inan and Hamid Shojanazeri and Han Zou and Hannah Wang and Hanwen Zha and Haroun Habeeb and Harrison Rudolph and Helen Suk and Henry Aspegren and Hunter Goldman and Hongyuan Zhan and Ibrahim Damlaj and Igor Molybog and Igor Tufanov and Ilias Leontiadis and Irina-Elena Veliche and Itai Gat and Jake Weissman and James Geboski and James Kohli and Janice Lam and Japhet Asher and Jean-Baptiste Gaya and Jeff Marcus and Jeff Tang and Jennifer Chan and Jenny Zhen and Jeremy Reizenstein and Jeremy Teboul and Jessica Zhong and Jian Jin and Jingyi Yang and Joe Cummings and Jon Carvill and Jon Shepard and Jonathan McPhie and Jonathan Torres and Josh Ginsburg and Junjie Wang and Kai Wu and Kam Hou U and Karan Saxena and Kartikay Khandelwal and Katayoun Zand and Kathy Matosich and Kaushik Veeraraghavan and Kelly Michelena and Keqian Li and Kiran Jagadeesh and Kun Huang and Kunal Chawla and Kyle Huang and Lailin Chen and Lakshya Garg and Lavender A and Leandro Silva and Lee Bell and Lei Zhang and Liangpeng Guo and Licheng Yu and Liron Moshkovich and Luca Wehrstedt and Madian Khabsa and Manav Avalani and Manish Bhatt and Martynas Mankus and Matan Hasson and Matthew Lennie and Matthias Reso and Maxim Groshev and Maxim Naumov and Maya Lathi and Meghan Keneally and Miao Liu and Michael L. Seltzer and Michal Valko and Michelle Restrepo and Mihir Patel and Mik Vyatskov and Mikayel Samvelyan and Mike Clark and Mike Macey and Mike Wang and Miquel Jubert Hermoso and Mo Metanat and Mohammad Rastegari and Munish Bansal and Nandhini Santhanam and Natascha Parks and Natasha White and Navyata Bawa and Nayan Singhal and Nick Egebo and Nicolas Usunier and Nikhil Mehta and Nikolay Pavlovich Laptev and Ning Dong and Norman Cheng and Oleg Chernoguz and Olivia Hart and Omkar Salpekar and Ozlem Kalinli and Parkin Kent and Parth Parekh and Paul Saab and Pavan Balaji and Pedro Rittner and Philip Bontrager and Pierre Roux and Piotr Dollar and Polina Zvyagina and Prashant Ratanchandani and Pritish Yuvraj and Qian Liang and Rachad Alao and Rachel Rodriguez and Rafi Ayub and Raghotham Murthy and Raghu Nayani and Rahul Mitra and Rangaprabhu Parthasarathy and Raymond Li and Rebekkah Hogan and Robin Battey and Rocky Wang and Russ Howes and Ruty Rinott and Sachin Mehta and Sachin Siby and Sai Jayesh Bondu and Samyak Datta and Sara Chugh and Sara Hunt and Sargun Dhillon and Sasha Sidorov and Satadru Pan and Saurabh Mahajan and Saurabh Verma and Seiji Yamamoto and Sharadh Ramaswamy and Shaun Lindsay and Shaun Lindsay and Sheng Feng and Shenghao Lin and Shengxin Cindy Zha and Shishir Patil and Shiva Shankar and Shuqiang Zhang and Shuqiang Zhang and Sinong Wang and Sneha Agarwal and Soji Sajuyigbe and Soumith Chintala and Stephanie Max and Stephen Chen and Steve Kehoe and Steve Satterfield and Sudarshan Govindaprasad and Sumit Gupta and Summer Deng and Sungmin Cho and Sunny Virk and Suraj Subramanian and Sy Choudhury and Sydney Goldman and Tal Remez and Tamar Glaser and Tamara Best and Thilo Koehler and Thomas Robinson and Tianhe Li and Tianjun Zhang and Tim Matthews and Timothy Chou and Tzook Shaked and Varun Vontimitta and Victoria Ajayi and Victoria Montanez and Vijai Mohan and Vinay Satish Kumar and Vishal Mangla and Vlad Ionescu and Vlad Poenaru and Vlad Tiberiu Mihailescu and Vladimir Ivanov and Wei Li and Wenchen Wang and Wenwen Jiang and Wes Bouaziz and Will Constable and Xiaocheng Tang and Xiaojian Wu and Xiaolan Wang and Xilun Wu and Xinbo Gao and Yaniv Kleinman and Yanjun Chen and Ye Hu and Ye Jia and Ye Qi and Yenda Li and Yilin Zhang and Ying Zhang and Yossi Adi and Youngjin Nam and Yu and Wang and Yu Zhao and Yuchen Hao and Yundi Qian and Yunlu Li and Yuzi He and Zach Rait and Zachary DeVito and Zef Rosnbrick and Zhaoduo Wen and Zhenyu Yang and Zhiwei Zhao and Zhiyu Ma},
      year={2024},
      howpublished={arXiv:2407.21783},
      url={https://arxiv.org/abs/2407.21783}, 
}

@inproceedings{vaswani2023attentionneed,
author = {Vaswani, Ashish and others},
FULLauthor = {Vaswani, Ashish and Shazeer, Noam and Parmar, Niki and Uszkoreit, Jakob and Jones, Llion and Gomez, Aidan N. and Kaiser, \L{}ukasz and Polosukhin, Illia},
title = {Attention is all you need},
year = {2017},
isbn = {9781510860964},
publisher = {Curran Associates Inc.},
address = {Red Hook, NY, USA},
booktitle = {Proceedings of the 31st International Conference on Neural Information Processing Systems},
pages = {6000–6010},
numpages = {11},
location = {Long Beach, California, USA},
series = {NIPS'17}
}

@article{wei2022emergentabilitieslargelanguage,
  author       = {Jason Wei and others},
  FULLauthor       = {Jason Wei and
                  Yi Tay and
                  Rishi Bommasani and
                  Colin Raffel and
                  Barret Zoph and
                  Sebastian Borgeaud and
                  Dani Yogatama and
                  Maarten Bosma and
                  Denny Zhou and
                  Donald Metzler and
                  Ed H. Chi and
                  Tatsunori Hashimoto and
                  Oriol Vinyals and
                  Percy Liang and
                  Jeff Dean and
                  William Fedus},
  title        = {Emergent Abilities of Large Language Models},
  journal      = {Transactions on Machine Learning Research},
  volume       = {2022},
  year         = {2022},
  OPTurl          = {https://openreview.net/forum?id=yzkSU5zdwD},
  bibsource    = {dblp computer science bibliography, https://dblp.org}
}

@InProceedings{llm_translation2023,
  title = 	 {Prompting Large Language Model for Machine Translation: A Case Study},
  author =       {Zhang, Biao and Haddow, Barry and Birch, Alexandra},
  booktitle = 	 {Proceedings of the 40th International Conference on Machine Learning},
  pages = 	 {41092--41110},
  year = 	 {2023},
  editor = 	 {Krause, Andreas and Brunskill, Emma and Cho, Kyunghyun and Engelhardt, Barbara and Sabato, Sivan and Scarlett, Jonathan},
  volume = 	 {202},
  series = 	 {Proceedings of Machine Learning Research},
  month = 	 {7},
  publisher =    {PMLR},
  OPTurl = 	 {https://proceedings.mlr.press/v202/zhang23m.html},
}

@misc{llm_coding2021,
      title={Evaluating Large Language Models Trained on Code}, 
      author={Mark Chen and others},
      FULLauthor={Mark Chen and Jerry Tworek and Heewoo Jun and Qiming Yuan and Henrique Ponde de Oliveira Pinto and Jared Kaplan and Harri Edwards and Yuri Burda and Nicholas Joseph and Greg Brockman and Alex Ray and Raul Puri and Gretchen Krueger and Michael Petrov and Heidy Khlaaf and Girish Sastry and Pamela Mishkin and Brooke Chan and Scott Gray and Nick Ryder and Mikhail Pavlov and Alethea Power and Lukasz Kaiser and Mohammad Bavarian and Clemens Winter and Philippe Tillet and Felipe Petroski Such and Dave Cummings and Matthias Plappert and Fotios Chantzis and Elizabeth Barnes and Ariel Herbert-Voss and William Hebgen Guss and Alex Nichol and Alex Paino and Nikolas Tezak and Jie Tang and Igor Babuschkin and Suchir Balaji and Shantanu Jain and William Saunders and Christopher Hesse and Andrew N. Carr and Jan Leike and Josh Achiam and Vedant Misra and Evan Morikawa and Alec Radford and Matthew Knight and Miles Brundage and Mira Murati and Katie Mayer and Peter Welinder and Bob McGrew and Dario Amodei and Sam McCandlish and Ilya Sutskever and Wojciech Zaremba},
      year={2021},
      eprint={2107.03374},
      archivePrefix={arXiv},
      primaryClass={cs.LG},
      url={https://arxiv.org/abs/2107.03374}, 
}

@article{dcr15,
    author = {Cohen, A. M. and Hersh, W. R. and Peterson, K. and Yen, Po-Yin},
    title = {Reducing Workload in Systematic Review Preparation Using Automated Citation Classification},
    journal = {Journal of the American Medical Informatics Association},
    volume = {13},
    number = {2},
    pages = {206-219},
    year = {2006},
    month = {03},
    issn = {1067-5027},
    OPTdoi = {10.1197/jamia.M1929},
    OPTeprint = {https://academic.oup.com/jamia/article-pdf/13/2/206/2178574/13-2-206.pdf},
}

@book{bird2009nltk,
  author    = {Steven Bird and Edward Loper and Ewan Klein},
  title     = {Natural Language Processing with Python},
  year      = {2009},
  publisher = {O’Reilly Media, Inc.},
}

@misc{llama3.18b_quant,
    author = "bartowski",
    title = "{Llamacpp imatrix Quantizations of Meta-Llama-3.1-8B-Instruct}",
    note = "Accessed: 2025-01-18",
url = {https://huggingface.co/bartowski/Meta-Llama-3.1-8B-Instruct-GGUF},
    year = "2024"
}

@misc{llama3.170b_quant,
    author       = "bartowski",
    title        = "{Llamacpp imatrix Quantizations of Meta-Llama-3.1-70B-Instruct}",
    note = "Accessed: 2025-01-18",
url = {https://huggingface.co/bartowski/Meta-Llama-3.1-70B-Instruct-GGUF},
    year         = "2024"
}

@misc{mistral7b_quant,
    author       = "bartowski",
    title        = "{Llamacpp imatrix Quantizations of Mistral-7B-Instruct-v0.3}",
    note = "Accessed: 2025-01-18",
url = {https://huggingface.co/bartowski/Mistral-7B-Instruct-v0.3-GGUF},
    year         = "2024"
}

@misc{gemma29b_quant,
    author       = "bartowski",
    title        = "{Llamacpp imatrix Quantizations of gemma-2-9b-it}",
    note = "Accessed: 2025-01-18",
url = {https://huggingface.co/bartowski/gemma-2-9b-it-GGUF},
    year         = "2024"
}

@inproceedings{mccallum98event,
  author={McCallum, A. and Nigam, K.},
  year={1998},
  title={A Comparison of Event Models for Naive {B}ayes Text Classification},
  booktitle={Proceedings of the {AAAI}/{ICML}-98 Workshop
             on Learning for Text Categorization},
  OPTpages={41--48},
  OPTvenue={Madison, WI, USA},
  publisher={AAAI Press},
  address={Menlo Park, CA, USA}}

@inproceedings{dumais98categorization,
  author={Dumais, S. T. and Platt, J. C. and Heckerman, D. and Sahami, M.},
  year={1998},
  title={Inductive Learning Algorithms and Representations for Text Categorization},
  booktitle={Proceedings of the Seventh International Conference
                     on Information and Knowledge Management},
  series={CIKM-98},
  pages={148--155},
  publisher={ACM Press}}

@inproceedings{joachims98categorization,
  author={Joachims, T.},
  year={1998},
  title={Text Categorization with Support Vector Machines:
         Learning with Many Relevant Features},
  booktitle={Proceedings of the Tenth European Conference on
             Machine Learning},
  series={ECML-98},
  OPTpages={137--142},
  OPTvenue={Chemnitz, Germany},
  publisher={Springer},
  OPTaddress={Berlin, Germany}}

@article{tong01svm,
  author={Tong, S. and Koller, D.},
  year={2001},
  title={Support Vector Machine Active Learning
         with Applications to Text Classification},
  journal={Journal of Machine Learning Research},
  volume={2},
  OPTnumber={1},
  pages={45--66}}

@article{sebastiani02categorization,
  author={Sebastiani, F.},
  year={2002},
  title={Machine Learning in Automated Text Categorization},
  journal={ACM Computing Surveys},
  volume={34},
  OPTnumber={1},
  pages = {1--47}}

@book{aggarwal12text,
  editor={Aggarwal, C. C. and Zhai, C.-X.},
  year={2012},
  title={Mining Text Data},
  publisher={Springer},
  address={New York, USA}}

@article{wu14forestexter,
  author={Wu, Q. and Ye, Y. and Zhang, H. and Ng, M. K. and Ho, S.-H.},
  year={2014},
  title={{ForesTexter}: An Efficient Random Forest Algorithm
          for Imbalanced Text Categorization},
  journal={Knowledge-Based Systems},
  volume={67},
  OPTnumber={Supplement C},
  pages={105--116}}

@inproceedings{wang20comparative,
  author={Wang, C. and Nulty, P. and Lillis, D.},
  year={2020},
  title={A Comparative Study on Word Embeddings in Deep Learning
         for Text Classification},
  booktitle={Proceedings of the Fourth International Conference on
             Natural Language Processing and Information Retrieval},
  series={NLPIR-2020},
  OPTpages={37--46},
  OPTvenue={Seoul, Korea},
  OPTpublisher={ACM Press}}

@inproceedings{jacobs22reducing,
  author={Jacobs, P. F. and Maillette de Buy Wenniger, G. and Wiering, M. and Schomaker, L.},
  year={2021},
  title={Active Learning for Reducing Labeling Effort in Text Classification Tasks},
  OPTeditor={Leiva, L. A. and Pruski, C. and Markovich, R. and Najjar, A. and Schommer, C.},
  booktitle={Artificial Intelligence and Machine Learning: Proceedings of the Thirty-Third Benelux on Artificial Intelligence},
  series={BNAIC/Benelearn-2021},
  publisher={Springer},
  OPTaddress={Cham, Switzerland},
  OPTvenue={Esch-sur-Alzette, Luxembourg}}

@article{cichosz23repclas,
  author={Cichosz, P.},
  year={2023},
  title={Bag of Words and Embedding Text Representation Methods
         for Medical Article Classification},
  journal={International Journal of Applied Mathematics and Computer Science},
  volume={33},
  OPTnumber={4},
  pages={603--621}}

@article{cichosz24repal,
  author={Cichosz, P.},
  year={2024},
  title={Active Learning for Medical Article Classification
         with Bag of Words and {FastText} Embeddings},
  journal={Applied Sciences},
  volume={14},
  OPTnumber={17},
  pages={7945},
  OPTdoi={10.3390/app14177945}}

@inproceedings{schroder24selftraining,
  title={Self-Training for Sample-Efficient Active Learning for Text Classification with Pre-Trained Language Models},
  author={Christopher Schr{\"o}der and Gerhard Heyer},
  booktitle={Conference on Empirical Methods in Natural Language Processing},
  year={2024},
  OPTurl={https://api.semanticscholar.org/CorpusID:270440282}
}

@inproceedings{devlin19bert,
  author={Devlin, J. and Chang, M.-W. and Lee, K. and Toutanova, K.},
  year={2019},
  title={{BERT}: Pre-Training of Deep Bidirectional Transformers
         for Language Understanding},
  booktitle={Proceedings of the Seventeenth Conference of the North American Chapter
             of the Association for Computational Linguistics:
             Human Language Technologies},
  series={NAACL-HLT-2019},
  OPTvenue={Minneapolis, MN, USA},
  pages={4171-4186},
  publisher={Association for Computational Linguistics},
  address={Stroudsburg, PA, USA}}

@article{breiman01rf,
  author={Breiman, L.},
  year={2001},
  title={Random Forests},
  journal={Machine Learning},
  volume={45},
  OPTnumber={1},
  pages={5--32}}

@misc{zhang2025summarization,
      title={A Comprehensive Survey on Process-Oriented Automatic Text Summarization with Exploration of {LLM}-Based Methods}, 
      author={Yang Zhang and Hanlei Jin and Dan Meng and Jun Wang and Jinghua Tan},
      year={2025},
      howpublished={arXiv:2403.02901},
      primaryClass={cs.AI},
      url={https://arxiv.org/abs/2403.02901}, 
}

@InProceedings{tan2023llmqa,
author="Tan, Yiming
and Min, Dehai
and Li, Yu
and Li, Wenbo
and Hu, Nan
and Chen, Yongrui
and Qi, Guilin",
editor="Payne, Terry R.
and Presutti, Valentina
and Qi, Guilin
and Poveda-Villal{\'o}n, Mar{\'i}a
and Stoilos, Giorgos
and Hollink, Laura
and Kaoudi, Zoi
and Cheng, Gong
and Li, Juanzi",
title="Can {ChatGPT} Replace Traditional {KBQA} Models? An In-Depth Analysis of the Question Answering Performance of the {GPT LLM} Family",
booktitle="The Semantic Web -- ISWC 2023",
year="2023",
publisher="Springer Nature Switzerland",
address="Cham",
pages="348--367",
isbn="978-3-031-47240-4"
}

@inproceedings{imani2023mathprompter,
  title={MathPrompter: Mathematical Reasoning using Large Language Models}, 
  author={Shima Imani and Liang Du and H. Shrivastava},
  booktitle={Annual Meeting of the Association for Computational Linguistics},
  year={2023},
  OPTurl={https://api.semanticscholar.org/CorpusID:257427208}
}

@inproceedings{wei2023chainofthoughtpromptingelicitsreasoning,
author = {Wei, Jason and Wang, Xuezhi and Schuurmans, Dale and Bosma, Maarten and Ichter, Brian and Xia, Fei and Chi, Ed H. and Le, Quoc V. and Zhou, Denny},
title = {Chain-of-Thought Prompting Elicits Reasoning in Large Language Models},
year = {2022},
isbn = {9781713871088},
publisher = {Curran Associates Inc.},
address = {Red Hook, NY, USA},
booktitle = {Proceedings of the 36th International Conference on Neural Information Processing Systems},
articleno = {1800},
numpages = {14},
location = {New Orleans, LA, USA},
series = {NIPS '22}
}

@inproceedings{mikolov13efficient,
  title={Efficient Estimation of Word Representations in Vector Space},
  author={Tomas Mikolov and Kai Chen and Gregory S. Corrado and Jeffrey Dean},
  booktitle={International Conference on Learning Representations},
  year={2013},
  OPTurl={https://api.semanticscholar.org/CorpusID:5959482}
}

@inproceedings{pennington14glove,
  author = {Pennington, J. and Socher, R. and Manning, C. D.},
  title = {{GloVe}: Global Vectors for Word Representation},
  year = {2014},
  booktitle = {Proceedings of the 2014 Conference on
               Empirical Methods in Natural Language Processing},
  series={EMNLP-2014},
  OPTpages={1532--1543},
  OPTvenue={Doha, Qatar},
  publisher={Association for Computational Linguistics},
  OPTaddress={Stroudsburg, PA, USA}}

@inproceedings{le14documents,
  author={Le, Q. V. and Mikolov, T.},
  year={2014},
  title={Distributed Representations of Sentences and Documents},
  booktitle={Proceedings of the Thirty-First International Conference
             on Machine Learning},
  series={ICML-2014},
  pages={1188--1196},
  OPTvenue={Beijing, China},
  publisher={JMLR Workshop and Conference Proceedings},
  OPTaddress={New York, USA}}

@article{bojanowski2016enriching,
  title={Enriching Word Vectors with Subword Information},
  author={Piotr Bojanowski and Edouard Grave and Armand Joulin and Tomas Mikolov},
  journal={Transactions of the Association for Computational Linguistics},
  year={2016},
  volume={5},
  pages={135-146},
  OPTurl={https://api.semanticscholar.org/CorpusID:207556454}
}

@inproceedings{akbik2018contextual,
  author={Akbik, A. and Blythe, D. and Vollgraf, R.},
  year={2018},
  title={Contextual String Embeddings for Sequence Labeling},
  booktitle = {Proceedings of the Twenty-Seventh International Conference
               on Computational Linguistics},
  series={COLING-2018},
  OPTvenue={Santa Fe, NM, USA},
  pages= {1638--1649},
  publisher={Association for Computational Linguistics},
  address={Stroudsburg, PA, USA}}

@inproceedings{davis06prroc,
  author = {Davis, J. and Goadrich, M.},
  title = {The Relationship between {P}recision-{R}ecall and {ROC} Curves},
  year = {2006},
  OPTisbn = {1595933832},
  publisher = {Association for Computing Machinery},
  OPTaddress = {New York, NY, USA},
  OPTurl = {https://doi.org/10.1145/1143844.1143874},
  OPTdoi = {10.1145/1143844.1143874},
  booktitle = {Proceedings of the 23rd International Conference on Machine Learning},
  pages = {233--240},
  OPTnumpages = {8},
  OPTlocation = {Pittsburgh, Pennsylvania, USA},
  series = {ICML-2006}
}

@inproceedings{yao2023treethoughtsdeliberateproblem,
title = "Tree of Thoughts: Deliberate Problem Solving with Large Language Models",
author = "Shunyu Yao and Dian Yu and Jeffrey Zhao and Izhak Shafran and Griffiths, \{Thomas L.\} and Yuan Cao and Karthik Narasimhan",
year = "2023",
language = "English (US)",
publisher = "Neural information processing systems foundation",
editor = "A. Oh and T. Neumann and A. Globerson and K. Saenko and M. Hardt and S. Levine",
booktitle = "Advances in Neural Information Processing Systems 36 - 37th Conference on Neural Information Processing Systems, NeurIPS 2023",

}

@inproceedings{reimers2019sentencebertsentenceembeddingsusing,
  author       = {Nils Reimers and
                  Iryna Gurevych},
  editor       = {Kentaro Inui and
                  Jing Jiang and
                  Vincent Ng and
                  Xiaojun Wan},
  title        = {Sentence-{BERT}: Sentence Embeddings using Siamese {BERT}-Networks},
  booktitle    = {Proceedings of the 2019 Conference on Empirical Methods in Natural
                  Language Processing and the 9th International Joint Conference on
                  Natural Language Processing, {EMNLP-IJCNLP} 2019, Hong Kong, China,
                  November 3-7, 2019},
  pages        = {3980--3990},
  publisher    = {Association for Computational Linguistics},
  year         = {2019},
  OPTurl          = {https://doi.org/10.18653/v1/D19-1410},
  OPTdoi          = {10.18653/V1/D19-1410},
  bibsource    = {dblp computer science bibliography, https://dblp.org}
}

@misc{wang2020minilmdeepselfattentiondistillation,
      title={MiniLM: Deep Self-Attention Distillation for Task-Agnostic Compression of Pre-Trained Transformers}, 
      author={Wenhui Wang and Furu Wei and Li Dong and Hangbo Bao and Nan Yang and Ming Zhou},
      year={2020},
      howpublished={arXiv:2002.10957},
      url={https://arxiv.org/abs/2002.10957}, 
}

@article{Sanh2019DistilBERTAD,
  title={Distil{BERT}, a distilled version of {BERT}: smaller, faster, cheaper and lighter},
  author={Victor Sanh and Lysandre Debut and Julien Chaumond and Thomas Wolf},
  howpublished={arXiv:1910.01108},
  year={2019},
  url={https://arxiv.org/abs/1910.01108}
}

@misc{song2020mpnetmaskedpermutedpretraining,
      title={{MPNet}: Masked and Permuted Pre-training for Language Understanding}, 
      author={Kaitao Song and Xu Tan and Tao Qin and Jianfeng Lu and Tie-Yan Liu},
      year={2020},
      howpublished={arXiv:2004.09297},
      url={https://arxiv.org/abs/2004.09297}, 
}

@inproceedings{gao2022simcsesimplecontrastivelearning,
    title = "{SimCSE}: Simple Contrastive Learning of Sentence Embeddings",
    author = "Gao, Tianyu  and
      Yao, Xingcheng  and
      Chen, Danqi",
    editor = "Moens, Marie-Francine  and
      Huang, Xuanjing  and
      Specia, Lucia  and
      Yih, Scott Wen-tau",
    booktitle = "Proceedings of the 2021 Conference on Empirical Methods in Natural Language Processing",
    month = nov,
    year = "2021",
    address = "Online and Punta Cana, Dominican Republic",
    publisher = "Association for Computational Linguistics",
    OPTurl = "https://aclanthology.org/2021.emnlp-main.552/",
    OPTdoi = "10.18653/v1/2021.emnlp-main.552",
    pages = "6894--6910"
}

@misc{openai2025models,
  author       = {{OpenAI}},
  title        = {{Models - OpenAI API}},
  year         = {2025},
  url = {https://platform.openai.com/docs/models},
  note         = {Accessed: 2025-08-06}
}

@misc{openai2025embeddings,
  author       = {{OpenAI}},
  title        = {{Vector embeddings - OpenAI API}},
  year         = {2025},
  url          = {https://platform.openai.com/docs/guides/embeddings},
  note         = {Accessed: 2025-11-21}
}

@misc{aliyun2025textembeddingv3,
  author       = {{Aliyun}},
  title        = {{Large Model Service Platform – Bailian Console}},
  year         = {2025},
  url = {https://bailian.console.aliyun.com/?tab=model#/model-market/detail/text-embedding-v3},
  note         = {Accessed: 2025-11-21}
}

@misc{openai2025dataprivacy,
  author       = {{OpenAI}},
  title        = {{Data controls in the OpenAI platform - OpenAI API}},
  year         = {2023},
  url = {https://platform.openai.com/docs/models},
  note         = {Accessed: 2025-10-09}
}

@article{scikit-learn,
  title={Scikit-learn: Machine Learning in {P}ython},
  author={Pedregosa, F. and others},
  FULLauthor={Pedregosa, F. and Varoquaux, G. and Gramfort, A. and Michel, V.
          and Thirion, B. and Grisel, O. and Blondel, M. and Prettenhofer, P.
          and Weiss, R. and Dubourg, V. and Vanderplas, J. and Passos, A. and
          Cournapeau, D. and Brucher, M. and Perrot, M. and Duchesnay, E.},
  journal={Journal of Machine Learning Research},
  volume={12},
  pages={2825--2830},
  year={2011}
}

@misc{llamacpppython2025docs,
  author       = {{llama-cpp-python contributors}},
  title        = {{llama-cpp-python Documentation}},
  year         = {2023},
  url = {https://llama-cpp-python.readthedocs.io/en/latest/},
  note         = {Accessed: 2025-08-06}
}

@misc{chase2025langchain,
  author       = {Harrison Chase},
  title        = {{LangChain}},
  year         = {2022},
  url = {https://github.com/langchain-ai/langchain},
  note         = {Accessed: 2025‑08‑06}
}

@article{gu2021pubmedbert,
author = {Gu, Yu and others},
FULLauthor = {Gu, Yu and Tinn, Robert and Cheng, Hao and Lucas, Michael and Usuyama, Naoto and Liu, Xiaodong and Naumann, Tristan and Gao, Jianfeng and Poon, Hoifung},
title = {Domain-Specific Language Model Pretraining for Biomedical Natural Language Processing},
year = {2021},
issue_date = {January 2022},
publisher = {Association for Computing Machinery},
address = {New York, NY, USA},
volume = {3},
number = {1},
OPTurl = {https://doi.org/10.1145/3458754},
OPTdoi = {10.1145/3458754},
journal = {ACM Trans. Comput. Healthcare},
OPTmonth = oct,
articleno = {2},
numpages = {23}
}

@inproceedings{zhang2025pushingthelimit,
author = {Zhang, Yazhou and Wang, Mengyao and Li, Qiuchi and Tiwari, Prayag and Qin, Jing},
title = {Pushing The Limit of {LLM} Capacity for Text Classification},
year = {2025},
isbn = {9798400713316},
publisher = {Association for Computing Machinery},
address = {New York, NY, USA},
OPTurl = {https://doi.org/10.1145/3701716.3715528},
OPTdoi = {10.1145/3701716.3715528},
booktitle = {Companion Proceedings of the ACM on Web Conference 2025},
pages = {1524–1528},
numpages = {5},
location = {Sydney NSW, Australia},
series = {WWW '25}
}

@misc{singh2025optimizingpromptrefinement,
  author       = {Singh, Juhi and Ao, Ziqiao and Antinome, Sebastian},
  title        = {Optimizing Prompt Refinement: Algorithmic Strategies for Large Language Model-Based Text Classification},
  howpublished={{SSRN} preprint},
  year         = {2025},
  url          = {https://ssrn.com/abstract=5525975},
  OPTdoi          = {10.2139/ssrn.5525975},
}

@article{shi2026anempiricalstudyofllms,
title = {An empirical study of {LLM}s via in-context learning for stance classification},
journal = {Information Processing \& Management},
volume = {63},
number = {1},
pages = {104322},
year = {2026},
issn = {0306-4573},
OPTdoi = {https://doi.org/10.1016/j.ipm.2025.104322},
OPTurl = {https://www.sciencedirect.com/science/article/pii/S0306457325002638},
author = {Lida Shi and Fausto Giunchiglia and Ran Luo and Daqian Shi and Rui Song and Xiaolei Diao and Hao Xu}
}

@misc{gemmateam2024gemma2improvingopen,
      title={Gemma 2: Improving Open Language Models at a Practical Size}, 
    author={{Gemma Team}},
      FULLauthor={Gemma Team and Morgane Riviere and Shreya Pathak and Pier Giuseppe Sessa and Cassidy Hardin and Surya Bhupatiraju and Léonard Hussenot and Thomas Mesnard and Bobak Shahriari and Alexandre Ramé and Johan Ferret and Peter Liu and Pouya Tafti and Abe Friesen and Michelle Casbon and Sabela Ramos and Ravin Kumar and Charline Le Lan and Sammy Jerome and Anton Tsitsulin and Nino Vieillard and Piotr Stanczyk and Sertan Girgin and Nikola Momchev and Matt Hoffman and Shantanu Thakoor and Jean-Bastien Grill and Behnam Neyshabur and Olivier Bachem and Alanna Walton and Aliaksei Severyn and Alicia Parrish and Aliya Ahmad and Allen Hutchison and Alvin Abdagic and Amanda Carl and Amy Shen and Andy Brock and Andy Coenen and Anthony Laforge and Antonia Paterson and Ben Bastian and Bilal Piot and Bo Wu and Brandon Royal and Charlie Chen and Chintu Kumar and Chris Perry and Chris Welty and Christopher A. Choquette-Choo and Danila Sinopalnikov and David Weinberger and Dimple Vijaykumar and Dominika Rogozińska and Dustin Herbison and Elisa Bandy and Emma Wang and Eric Noland and Erica Moreira and Evan Senter and Evgenii Eltyshev and Francesco Visin and Gabriel Rasskin and Gary Wei and Glenn Cameron and Gus Martins and Hadi Hashemi and Hanna Klimczak-Plucińska and Harleen Batra and Harsh Dhand and Ivan Nardini and Jacinda Mein and Jack Zhou and James Svensson and Jeff Stanway and Jetha Chan and Jin Peng Zhou and Joana Carrasqueira and Joana Iljazi and Jocelyn Becker and Joe Fernandez and Joost van Amersfoort and Josh Gordon and Josh Lipschultz and Josh Newlan and Ju-yeong Ji and Kareem Mohamed and Kartikeya Badola and Kat Black and Katie Millican and Keelin McDonell and Kelvin Nguyen and Kiranbir Sodhia and Kish Greene and Lars Lowe Sjoesund and Lauren Usui and Laurent Sifre and Lena Heuermann and Leticia Lago and Lilly McNealus and Livio Baldini Soares and Logan Kilpatrick and Lucas Dixon and Luciano Martins and Machel Reid and Manvinder Singh and Mark Iverson and Martin Görner and Mat Velloso and Mateo Wirth and Matt Davidow and Matt Miller and Matthew Rahtz and Matthew Watson and Meg Risdal and Mehran Kazemi and Michael Moynihan and Ming Zhang and Minsuk Kahng and Minwoo Park and Mofi Rahman and Mohit Khatwani and Natalie Dao and Nenshad Bardoliwalla and Nesh Devanathan and Neta Dumai and Nilay Chauhan and Oscar Wahltinez and Pankil Botarda and Parker Barnes and Paul Barham and Paul Michel and Pengchong Jin and Petko Georgiev and Phil Culliton and Pradeep Kuppala and Ramona Comanescu and Ramona Merhej and Reena Jana and Reza Ardeshir Rokni and Rishabh Agarwal and Ryan Mullins and Samaneh Saadat and Sara Mc Carthy and Sarah Cogan and Sarah Perrin and Sébastien M. R. Arnold and Sebastian Krause and Shengyang Dai and Shruti Garg and Shruti Sheth and Sue Ronstrom and Susan Chan and Timothy Jordan and Ting Yu and Tom Eccles and Tom Hennigan and Tomas Kocisky and Tulsee Doshi and Vihan Jain and Vikas Yadav and Vilobh Meshram and Vishal Dharmadhikari and Warren Barkley and Wei Wei and Wenming Ye and Woohyun Han and Woosuk Kwon and Xiang Xu and Zhe Shen and Zhitao Gong and Zichuan Wei and Victor Cotruta and Phoebe Kirk and Anand Rao and Minh Giang and Ludovic Peran and Tris Warkentin and Eli Collins and Joelle Barral and Zoubin Ghahramani and Raia Hadsell and D. Sculley and Jeanine Banks and Anca Dragan and Slav Petrov and Oriol Vinyals and Jeff Dean and Demis Hassabis and Koray Kavukcuoglu and Clement Farabet and Elena Buchatskaya and Sebastian Borgeaud and Noah Fiedel and Armand Joulin and Kathleen Kenealy and Robert Dadashi and Alek Andreev},
      year={2024},
      howpublished={arXiv:2408.00118},
      url={https://arxiv.org/abs/2408.00118}, 
}

@inproceedings{
    he2021deberta,
    title={DeBERTa: Decoding-enhanced BERT with Disentangled Attention},
    author={Pengcheng He and Xiaodong Liu and Jianfeng Gao and Weizhu Chen},
    booktitle={International Conference on Learning Representations},
    year={2021},
    OPTurl={https://openreview.net/forum?id=XPZIaotutsD}
}

@article{medibiodeberta,
    author = {Kim, Eunhui and Jeong, Yuna and Choi, Myung-seok},
    year = {2023},
    month = {01},
    pages = {1-1},
    title = {{MediBioDeBERTa}: Biomedical Language Model with Continuous Learning and Intermediate Fine-Tuning},
    volume = {PP},
    journal = {IEEE Access},
    OPTdoi = {10.1109/ACCESS.2023.3341612}
}

@inproceedings{huggingfacetransformers,
    author = {{Wolf, Thomas et al.}},
    FULLauthor = {Wolf, Thomas and Debut, Lysandre and Sanh, Victor and Chaumond, Julien and Delangue, Clement and Moi, Anthony and Cistac, Pierric and Rault, Tim and Louf, Remi and Funtowicz, Morgan and Davison, Joe and Shleifer, Sam and Platen, Patrick and Ma, Clara and Jernite, Yacine and Plu, Julien and Xu, Canwen and Scao, Teven and Gugger, Sylvain and Rush, Alexander},
    year = {2020},
    month = {01},
    pages = {38-45},
    title = {Transformers: State-of-the-Art Natural Language Processing},
    OPTdoi = {10.18653/v1/2020.emnlp-demos.6}
}

@misc{liu2019roberta,
      title={{RoBERTa}: A Robustly Optimized {BERT} Pretraining Approach}, 
      author={Yinhan Liu and Myle Ott and Naman Goyal and Jingfei Du and Mandar Joshi and Danqi Chen and Omer Levy and Mike Lewis and Luke Zettlemoyer and Veselin Stoyanov},
      year={2019},
      howpublished={arXiv:1907.11692},
      url={https://arxiv.org/abs/1907.11692}, 
}

\newpage
\appendix

\section{Prompts}\label{app:prompts}

Table~\ref{tab:prompts} contains prompts used in the Prompts experiment.

Table~\ref{tab:prompts2} contains prompts used in the Output experiment, as well as in the Baseline experiment. For the Word output processing method, prompts from the previous experiment were used directly. For the Token-Word and Token-JSON output processing methods, the same prompts were used as for the Word and JSON output processing methods, except that for Token-JSON an assistant message containing \texttt{\{"high\_quality":} for prompt B or \texttt{\{"included":} for prompt F was added.

Table~\ref{tab:prompts3} contains the prompt used in Few-Shot-Count and Few-Shot-Selection experiments, as well as in the Baseline experiment. The first user message and the assistant message were repeated for each example.

Table~\ref{tab:prompts4} contains the prompts used in the Advanced-Techniques experiment.

The tables contain tags with the following meanings:
\begin{itemize}
    \item \{text\} -- the place to insert the classified sample,
    \item \{example\_text\} -- the place to insert the example,
    \item \{example\_answer\} -- the place to insert the correct answer for the example,
    \item \{topic\} -- the place to insert the review topic corresponding to the dataset,
    \item \{last\_gen\} -- for methods with multiple LLM requests, the place to insert the output of the previous model request; for evaluating models with the tree of thoughts method, this tag means the place where the evaluated generation was inserted.
\end{itemize}

{\small\tabcolsep=3pt
\begin{longtable}{| c | c | m{11cm} |}
    \caption{Prompts used in Prompts experiment.} \\
    \hline
    Name & Role & Message content \\ \hline\hline \endfirsthead

    \endfoot
    \hline \endlastfoot

    \multirow{2}{*}{A} & system     & {Is this an abstract of a high-quality article on \{topic\}? Answer yes or no.} \\\cline{2-3}
                           & user & \{text\} \\\hline
    \multirow{2}{*}[-16pt]{B} & system     & You are a helpful assistant. \\\cline{2-3}
                           & user & {Is this an abstract of a high-quality article on \{topic\}? Answer yes or no.\newline "\{text\}"} \\\hline
    \multirow{2}{*}{C} & system     & {You are an expert on medical articles classification. You are going to help classify article abstracts. The user will pass in a text and you should answer yes if it is an abstract of a high-quality article on the topic of \{topic\} and no otherwise.} \\\cline{2-3}
                           & user & \{text\} \\\hline
    \multirow{2}{*}[-8pt]{D} & system     & {You are an expert on medical articles classification. You are going to help classify article abstracts. Answer with a single word only, yes or no.} \\\cline{2-3}
                           & user & {Is the following an abstract of a high-quality article on the topic of \{topic\}? "\{text\}"} \\\hline
    \multirow{2}{*}{E} & system     & {You are an AI assistant and you are very good at text classification. You are going to help classify medical article abstracts. You will be given an abstract of an article considered for inclusion in a review of scientific articles on the topic of \{topic\}. You should reply yes if the article should be included, and no otherwise.} \\\cline{2-3}
                           & user & \{text\} \\\hline
    \multirow{2}{*}[-12pt]{F} & system     & {You are an AI assistant and you are very good at text classification. You are going to help classify medical article abstracts. You will be given an abstract of an article considered for inclusion in a review of scientific articles on the topic of \{topic\}.} \\\cline{2-3}
                           & user & {Should the article the following abstract describes be included in the review on \{topic\}? "\{text\}" Reply with a single word, yes or no.}
    \label{tab:prompts}
\end{longtable}
}

{\small\tabcolsep=3pt
\begin{longtable}{| c | c | m{10cm} |}
    \caption{Prompts used in Output Experiment.} \\
    \hline
    Name & Role & Message content \\ \hline\hline \endfirsthead

    \endfoot
    \hline \endlastfoot

    \multirow{2}{*}[-20pt]{B (Score)} & system     & {You are a helpful assistant.} \\\cline{2-3}
                                & user & {Is this an abstract of a high-quality article on \{topic\}? Rank the article on a scale from 0 to 5, where 5 means that it is a high-quality article on \{topic\}.\newline
                                "\{text\}"} \\\hline
    \multirow{2}{*}[-20pt]{F (Score)} & system     & {You are an AI assistant and you are very good at text classification. You are going to help classify medical article abstracts. You will be given an abstract of an article considered for inclusion in a review of scientific articles on the topic of \{topic\}.} \\\cline{2-3}
                                & user & {Should the article the following abstract describes be included in the review on \{topic\}? "\{text\}" Rank the article on a scale from 0 to 5, where 0 means "should not be included" and 5 means "should definitely be included".} \\\hline
                            \pagebreak\hline
    \multirow{2}{*}[-20pt]{B (JSON)} & system     & {You are a helpful assistant.} \\\cline{2-3}
                                & user & {Is this an abstract of a high-quality article on \{topic\}? Answer in JSON format: \{"high\_quality": false\} or \{"high\_quality": true\}. \newline "\{text\}"} \\\hline
    \multirow{2}{*}[-14pt]{F (JSON)} & system     & {You are an AI assistant and you are very good at text classification. You are going to help classify medical article abstracts. You will be given an abstract of an article considered for inclusion in a review of scientific articles on the topic of {topic}.} \\\cline{2-3}
                                & user & {Should the article the following abstract describes be included in the review on \{topic\}? "\{text\}" Answer in JSON format: \{"included": false\} or \{"included": true\}.}
    \label{tab:prompts2}
\end{longtable}
}

{\small\tabcolsep=3pt
\begin{longtable}{| c | c | m{10cm} |}
    \caption{Prompt used in Few-Shot-Count and Few-Shot-Selection experiments.} \\
    \hline
    Name & Role & Message content \\ \hline\hline \endfirsthead

    \endfoot
    \hline \endlastfoot

    \multirow{4}{*}[-25pt]{B (few-shot)} & system     & {You are a helpful assistant.} \\\cline{2-3}
                           & user &           {Is this an abstract of a high-quality article on \{topic\}? Answer yes or no.\newline "\{example\_text\}"} \\\cline{2-3}
                           & assistant & \{example\_answer\} \\\cline{2-3}
                           & user & {Is this an abstract of a high-quality article on \{topic\}? Answer yes or no.\newline "\{text\}"}
    \label{tab:prompts3}
\end{longtable}
}

{\small\tabcolsep=3pt
\begin{longtable}{| c | c | m{10cm} |}
    \caption{Prompts used in Advanced-Techniques experiment.} \\
    \hline
    Name & Role & Message content \\ \hline\hline \endfirsthead

    \endfoot
    \hline \endlastfoot

    \multirow{3}{*}[-12pt]{CoT (1 call)} & system & You are an AI assistant and you are very good at text classification. You are going to help classify medical article abstracts. You will be given an abstract of an article considered for inclusion in a review of scientific articles on the topic of \{topic\}. \\\cline{2-3}
                           & user & Should the article the following abstract describes be included in the review on \{topic\}? "\{text\}" Answer in JSON format: \{"included": false\} or \{"included": true\}. \\\cline{2-3}
                            & assistant & Let's think step by step. \\\hline
    \multirow{3}{*}[-15pt]{\makecell{CoT (2 calls) \\ Model $1$}} & system & You are an AI assistant and you are very good at text classification. You are going to help classify medical article abstracts. You will be given an abstract of an article considered for inclusion in a review of scientific articles on the topic of \{topic\}. \\\cline{2-3}
                           & user & From this abstract, extract phrases that could point towards the article containing high quality evidence or a lack thereof. "\{text\}" Give your answer as a JSON list. \\\cline{2-3}
                            & assistant & [" \\\hline
    \multirow{3}{*}[-46pt]{\makecell{CoT (2 calls) \\ Model $2$}} & system & You are an AI assistant and you are very good at text classification. You are going to help classify medical article abstracts. You will be given an abstract of an article considered for inclusion in a review of scientific articles on the topic of \{topic\}, together with some phrases extracted from it that point towards it contining high quality evidence or not. \\\cline{2-3}
     & user & {Should the article the following abstract describes be included in the review on \{topic\}? "\{text\}" \newline\newline
                                                       Here are some phrases extracted from it in JSON format: ["\{last\_gen\} \newline\newline
                                                       Answer in JSON format: \{"included": false\} or \{"included": true\}.} \\\cline{2-3}
                            & assistant & \{"included": " \\\hline
    \multirow{4}{*}[-26pt]{Chunking} & system & {You are an AI assistant and you are very good at text classification. You are going to help classify medical article abstracts. You will be given a fragment of an abstract of an article considered for inclusion in a review of scientific articles on the topic of \{topic\}.} \\\cline{2-3}
                           & user & {Should the article the following abstract fragment describes be included in the review on \{topic\}? "\{example\_text\}" Reply with a single word, yes or no.} \\\cline{2-3}
                            & assistant & \{example\_answer\} \\\cline{2-3}
                           & user & {Should the article the following abstract fragment describes be included in the review on \{topic\}? "\{text\}" Reply with a single word, yes or no.} \\\hline
                            \pagebreak\hline
    \multirow{3}{*}[-16pt]{\makecell{ToT \\ Model $1$ \\ for generation}}
                               & system & {You are an AI assistant and you are very good at natural language processing. You are going to help analyze medical article abstracts. You will be given an abstract of an article considered for inclusion in a review of scientific articles on the topic of \{topic\}.} \\\cline{2-3}
                           & user & {From this abstract, extract phrases that could point towards the article containing high quality evidence or a lack thereof. "\{text\}" Your final answer should be a JSON list of the extracted phrases.} \\\cline{2-3}
                            & assistant & Let's think step by step. \\\hline
    \multirow{3}{*}[-40pt]{\makecell{ToT \\ Model $2$ \\ for generation}}
                               & system & {You are an AI assistant and you are very good at text classification. You are going to help classify medical article abstracts. You will be given an abstract of an article considered for inclusion in a review of scientific articles on the topic of \{topic\}, together with some phrases extracted from it that point towards it contining high quality evidence or not.} \\\cline{2-3}
                           & user & {Should the article the following abstract describes be included in the review on \{topic\}? "\{text\}" \newline\newline Here are some phrases extracted from it in JSON format: ["\{last\_gen\} \newline\newline Give your final answer in JSON format: \{"included": false\} or \{"included": true\}.} \\\cline{2-3}
                            & assistant & Let's think step by step. \\\hline
    \multirow{3}{*}[-50pt]{\makecell{ToT \\ Model $1$ \\ for evaluation}}
                               & system & {You are an AI assistant and you an expert at natural language processing. You are going to help analyze medical article abstracts. You will be given an abstract of an article considered for inclusion in a review of scientific articles on the topic of \{topic\}.} \\\cline{2-3}
                           & user & {"\{text\}"\newline\newline The following are phrases that could point to the article containing high-quality evidence in JSON format extracted from the above text: \{last\_gen\} \newline\newline Rate the quality of extraction from 0 to 5. Answer in JSON format: \{"quality": n\} where n is a number from 0 to 5.} \\\cline{2-3}
                            & assistant & \{"quality":  \\\hline
    \multirow{3}{*}[-45pt]{\makecell{ToT \\ Model $2$ \\ for evaluation}}
                               & system & {You are an AI assistant and you an expert at natural language processing. You are going to help analyze medical article abstracts. You will be given an abstract of an article considered for inclusion in a review of scientific articles on the topic of \{topic\}.} \\\cline{2-3}
                           & user & {For the abstract: "\{text\}" the following was an answer to a question whether it should be included in a review of articles on the topic of \{topic\}.\newline "\{last\_gen\}" \newline\newline Rate the quality of the answer from 0 to 5. Answer in JSON format: \{"quality": n\} where n is a number from 0 to 5.} \\\cline{2-3}
                            & assistant & \{"quality":
    \label{tab:prompts4}
\end{longtable}
}

\section{MCC results over datasets}\label{app:mcc}

The following tables contain the detailed MCC scores for each of the variants and datasets experimented with in Output, Few-Shot-Count, Few-Shot-Selection and Baseline experiments, which were omitted from the main manuscript.

\begin{table}[!htb]
\caption{Macro-averaged MCC scores for the Output experiment.}\label{tab:dcr.output.mcc}
\resizebox{\linewidth}{!}{%
\centering
\begin{tabular}{|c|c|c|c|c|c|}
\hline
Output & Word & JSON & Score & Token-Word & \textbf{Token-JSON} \\
\hline\hline
ACEInhibitors & $0.099$ & $0.091$ & $0.079$ & $\mathbf{0.099}$ & $\mathbf{0.099}$ \\
\hline
ADHD & $0.482$ & $0.504$ & $0.419$ & $0.510$ & $\mathbf{0.516}$ \\
\hline
Antihistamines & $0.234$ & $0.228$ & $0.168$ & $\mathbf{0.240}$ & $0.221$ \\
\hline
AtypicalAntipsychotics & $0.145$ & $\mathbf{0.176}$ & $0.097$ & $0.151$ & $0.168$ \\
\hline
BetaBlockers & $0.130$ & $0.122$ & $0.106$ & $\mathbf{0.136}$ & $0.135$ \\
\hline
CalciumChannelBlockers & $0.122$ & $0.114$ & $0.110$ & $\mathbf{0.126}$ & $0.106$ \\
\hline
Estrogens & $0.178$ & $0.197$ & $0.106$ & $0.192$ & $\mathbf{0.219}$ \\
\hline
NSAIDS & $0.106$ & $0.149$ & $0.084$ & $0.125$ & $\mathbf{0.150}$ \\
\hline
Opiods & $0.054$ & $0.058$ & $0.041$ & $0.064$ & $\mathbf{0.079}$ \\
\hline
OralHypoglycemics & $0.135$ & $0.146$ & $0.082$ & $0.136$ & $\mathbf{0.146}$ \\
\hline
ProtonPumpInhibitors & $0.125$ & $\mathbf{0.152}$ & $0.075$ & $0.135$ & $0.140$ \\
\hline
SkeletalMuscleRelaxants & $0.170$ & $0.153$ & $0.140$ & $\mathbf{0.173}$ & $0.144$ \\
\hline
Statins & $0.107$ & $0.111$ & $0.092$ & $0.117$ & $\mathbf{0.119}$ \\
\hline
Triptans & $0.295$ & $\mathbf{0.341}$ & $0.232$ & $0.314$ & $0.320$ \\
\hline
UrinaryIncontinence & $0.072$ & $0.093$ & $0.068$ & $0.076$ & $\mathbf{0.094}$ \\
\hline
\end{tabular}
}
\end{table}

\begin{table}[!htb]
\caption{Macro-averaged MCC scores the Few-Shot-Count experiment.}\label{tab:dcr.n_shot.mcc}
\centering
\begin{tabular}{|c|c|c|c|c|c|}
\hline
Examples & 0 & \textbf{1} & 2 & 3 & 4 \\
\hline\hline
ACEInhibitors & $0.099$ & $\mathbf{0.116}$ & $0.113$ & $0.091$ & $0.088$ \\
\hline
ADHD & $0.470$ & $\mathbf{0.482}$ & $0.480$ & $0.468$ & $0.466$ \\
\hline
Antihistamines & $\mathbf{0.214}$ & $0.202$ & $0.184$ & $0.181$ & $0.171$ \\
\hline
AtypicalAntipsychotics & $0.153$ & $\mathbf{0.182}$ & $0.153$ & $0.142$ & $0.137$ \\
\hline
BetaBlockers & $0.129$ & $\mathbf{0.153}$ & $0.126$ & $0.125$ & $0.123$ \\
\hline
CalciumChannelBlockers & $0.134$ & $\mathbf{0.174}$ & $0.137$ & $0.116$ & $0.124$ \\
\hline
Estrogens & $0.194$ & $\mathbf{0.201}$ & $0.165$ & $0.138$ & $0.140$ \\
\hline
NSAIDS & $0.138$ & $\mathbf{0.139}$ & $0.109$ & $0.118$ & $0.112$ \\
\hline
Opiods & $0.062$ & $\mathbf{0.066}$ & $0.051$ & $0.049$ & $0.049$ \\
\hline
OralHypoglycemics & $0.136$ & $\mathbf{0.178}$ & $0.147$ & $0.143$ & $0.146$ \\
\hline
ProtonPumpInhibitors & $0.136$ & $\mathbf{0.161}$ & $0.142$ & $0.115$ & $0.108$ \\
\hline
SkeletalMuscleRelaxants & $0.183$ & $\mathbf{0.231}$ & $0.189$ & $0.188$ & $0.190$ \\
\hline
Statins & $0.110$ & $\mathbf{0.123}$ & $0.114$ & $0.117$ & $0.115$ \\
\hline
Triptans & $0.313$ & $\mathbf{0.324}$ & $0.290$ & $0.283$ & $0.290$ \\
\hline
UrinaryIncontinence & $0.098$ & $\mathbf{0.168}$ & $0.160$ & $0.146$ & $0.148$ \\
\hline
\end{tabular}
\end{table}

\begin{table}[!htb]
\caption{Macro-averaged MCC scores for the Few-Shot-Selection experiment.}\label{tab:dcr.select.mcc}
\resizebox{\linewidth}{!}{%
\centering
\begin{tabular}{|c|c|c|c|c|c|}
\hline
Method & \textbf{random} & minilm & distroberta & \textbf{mpnet} & mpnet-bal \\
\hline\hline
ACEInhibitors & $0.102$ & $0.091$ & $\mathbf{0.104}$ & $0.098$ & $0.077$ \\
\hline
ADHD & $0.482$ & $0.468$ & $0.472$ & $\mathbf{0.487}$ & $0.372$ \\
\hline
Antihistamines & $\mathbf{0.197}$ & $0.180$ & $0.190$ & $0.191$ & $0.157$ \\
\hline
AtypicalAntipsychotics & $\mathbf{0.156}$ & $0.142$ & $0.144$ & $0.138$ & $0.118$ \\
\hline
BetaBlockers & $\mathbf{0.130}$ & $0.125$ & $\mathbf{0.130}$ & $0.125$ & $0.120$ \\
\hline
CalciumChannelBlockers & $0.112$ & $0.115$ & $0.121$ & $\mathbf{0.128}$ & $0.126$ \\
\hline
Estrogens & $0.160$ & $0.138$ & $0.154$ & $\mathbf{0.163}$ & $0.141$ \\
\hline
NSAIDS & $\mathbf{0.139}$ & $0.118$ & $0.112$ & $0.118$ & $0.092$ \\
\hline
Opiods & $0.052$ & $0.049$ & $0.049$ & $\mathbf{0.054}$ & $0.028$ \\
\hline
OralHypoglycemics & $\mathbf{0.157}$ & $0.145$ & $0.144$ & $0.152$ & $0.129$ \\
\hline
ProtonPumpInhibitors & $0.115$ & $0.115$ & $0.121$ & $\mathbf{0.132}$ & $0.101$ \\
\hline
SkeletalMuscleRelaxants & $0.146$ & $\mathbf{0.188}$ & $0.181$ & $0.187$ & $0.154$ \\
\hline
Statins & $\mathbf{0.121}$ & $0.117$ & $0.117$ & $0.120$ & $0.105$ \\
\hline
Triptans & $\mathbf{0.318}$ & $0.285$ & $0.274$ & $0.289$ & $0.271$ \\
\hline
UrinaryIncontinence & $\mathbf{0.152}$ & $0.150$ & $0.131$ & $0.150$ & $0.126$ \\
\hline
\end{tabular}
}
\end{table}

\begin{table}[!htb]
\caption{Macro-averaged MCC scores for the Baseline experiment.}\label{tab:dcr.final.mcc}
\resizebox{\linewidth}{!}{%
\centering
\begin{tabular}{|c|c|c|c|c|c|c|c|c|c|c|c|}
\hline
Alg & gemm & ll-70b & ll-8b & mistr & gptmi & gptna & \textbf{nb} & rf-d & rf-t & debe & scid \\
\hline\hline
Ace & $0.164$ & $0.134$ & $0.121$ & $-0.003$ & $0.135$ & $0.190$ & $0.327$ & $0.219$ & $0.243$ & $\mathbf{0.368}$ & $0.360$ \\
\hline
Adh & $0.634$ & $\mathbf{0.646}$ & $0.563$ & $0.044$ & $0.570$ & $0.587$ & $0.558$ & $0.593$ & $0.562$ & $0.534$ & $0.599$ \\
\hline
Ant & $0.261$ & $0.269$ & $0.193$ & $0.036$ & $0.223$ & $0.277$ & $0.356$ & $0.268$ & $\mathbf{0.417}$ & $0.192$ & $0.038$ \\
\hline
Aty & $0.219$ & $0.176$ & $0.135$ & $0.047$ & $0.228$ & $0.312$ & $0.421$ & $0.385$ & $0.270$ & $0.378$ & $\mathbf{0.473}$ \\
\hline
Bet & $0.146$ & $0.164$ & $0.162$ & $0.046$ & $0.137$ & $0.207$ & $0.304$ & $0.232$ & $0.205$ & $0.237$ & $\mathbf{0.350}$ \\
\hline
Cal & $0.140$ & $0.146$ & $0.168$ & $0.028$ & $0.163$ & $0.187$ & $0.478$ & $0.507$ & $0.509$ & $0.318$ & $\mathbf{0.518}$ \\
\hline
Est & $0.233$ & $0.202$ & $0.166$ & $0.013$ & $0.255$ & $0.336$ & $\mathbf{0.485}$ & $0.450$ & $0.385$ & $0.213$ & $0.341$ \\
\hline
Nsa & $0.174$ & $0.112$ & $0.129$ & $0.032$ & $0.143$ & $0.336$ & $\mathbf{0.573}$ & $0.531$ & $0.511$ & $0.461$ & $0.568$ \\
\hline
Opi & $0.046$ & $0.061$ & $0.074$ & $0.013$ & $0.050$ & $0.098$ & $0.155$ & $0.147$ & $0.144$ & $0.259$ & $\mathbf{0.338}$ \\
\hline
Ora & $0.211$ & $0.197$ & $0.142$ & $0.026$ & $0.231$ & $0.301$ & $0.367$ & $0.329$ & $0.363$ & $0.350$ & $\mathbf{0.395}$ \\
\hline
Pro & $0.193$ & $0.128$ & $0.114$ & $0.048$ & $0.201$ & $0.303$ & $\mathbf{0.363}$ & $0.314$ & $0.314$ & $0.279$ & $0.353$ \\
\hline
Ske & $0.229$ & $0.244$ & $0.223$ & $0.026$ & $0.221$ & $0.201$ & $\mathbf{0.365}$ & $0.155$ & $0.143$ & $0.040$ & $0.236$ \\
\hline
Sta & $0.157$ & $0.138$ & $0.138$ & $0.034$ & $0.165$ & $0.200$ & $\mathbf{0.294}$ & $0.181$ & $0.164$ & $0.164$ & $0.256$ \\
\hline
Tri & $0.395$ & $0.344$ & $0.299$ & $0.058$ & $0.361$ & $0.470$ & $\mathbf{0.594}$ & $0.580$ & $0.555$ & $0.551$ & $0.590$ \\
\hline
Uri & $0.192$ & $0.160$ & $0.200$ & $-0.028$ & $0.233$ & $0.363$ & $\mathbf{0.552}$ & $0.524$ & $0.514$ & $0.503$ & $0.520$ \\
\hline
\end{tabular}
}
\end{table}

\clearpage

\section{Detailed results over models}\label{app:models}

The following tables contain detailed scores for each of the variants and models experimented with in all the experiments, averaged over the datasets, which were omitted from the main manuscript.

\begin{table}[!htb]
\caption{Performance scores for the Prompts experiment macro-averaged over all datasets.}\label{tab:dcr.prompt.avg}
\centering
\begin{tabular}{|c|c|c|c|c|c|c|}
\hline
Prompt & Model & Accuracy & Recall & Precision & F1 Score & MCC \\
\hline\hline
\multirow{5}{*}{A} & Gemma 2 9B & $0.412$ & $0.933$ & $0.231$ & $0.356$ & $0.175$ \\
\cline{2-7}
 & Llama 3.1 70B & $0.444$ & $0.921$ & $0.245$ & $0.369$ & $0.196$ \\
\cline{2-7}
 & Llama 3.1 8B & $0.347$ & $0.950$ & $0.213$ & $0.334$ & $0.124$ \\
\cline{2-7}
 & Mistral 7B & $0.204$ & $\mathbf{0.996}$ & $0.189$ & $0.303$ & $0.030$ \\
\cline{2-7}
 & Average & $0.352$ & $0.950$ & $0.219$ & $0.341$ & $0.131$ \\
\hline
\hline
\multirow{5}{*}{B} & Gemma 2 9B & $0.440$ & $0.916$ & $0.242$ & $0.367$ & $0.193$ \\
\cline{2-7}
 & Llama 3.1 70B & $\mathbf{0.520}$ & $0.869$ & $\mathbf{0.266}$ & $\mathbf{0.389}$ & $\mathbf{0.230}$ \\
\cline{2-7}
 & Llama 3.1 8B & $0.397$ & $0.951$ & $0.228$ & $0.353$ & $0.172$ \\
\cline{2-7}
 & Mistral 7B & $0.225$ & $0.993$ & $0.191$ & $0.307$ & $0.057$ \\
\cline{2-7}
 & Average & $\mathbf{0.396}$ & $0.932$ & $\mathbf{0.232}$ & $\mathbf{0.354}$ & $0.163$ \\
\hline
\hline
\multirow{5}{*}{C} & Gemma 2 9B & $0.400$ & $0.947$ & $0.227$ & $0.352$ & $0.167$ \\
\cline{2-7}
 & Llama 3.1 70B & $0.445$ & $0.923$ & $0.245$ & $0.369$ & $0.193$ \\
\cline{2-7}
 & Llama 3.1 8B & $0.310$ & $0.986$ & $0.210$ & $0.332$ & $0.124$ \\
\cline{2-7}
 & Mistral 7B & $0.219$ & $0.994$ & $0.191$ & $0.306$ & $0.042$ \\
\cline{2-7}
 & Average & $0.344$ & $\mathbf{0.962}$ & $0.218$ & $0.340$ & $0.131$ \\
\hline
\hline
\multirow{5}{*}{D} & Gemma 2 9B & $0.416$ & $0.931$ & $0.232$ & $0.357$ & $0.173$ \\
\cline{2-7}
 & Llama 3.1 70B & $0.471$ & $0.909$ & $0.252$ & $0.376$ & $0.207$ \\
\cline{2-7}
 & Llama 3.1 8B & $0.395$ & $0.948$ & $0.230$ & $0.354$ & $0.168$ \\
\cline{2-7}
 & Mistral 7B & $0.249$ & $0.986$ & $0.193$ & $0.310$ & $0.063$ \\
\cline{2-7}
 & Average & $0.383$ & $0.943$ & $0.227$ & $0.349$ & $0.153$ \\
\hline
\hline
\multirow{5}{*}{E} & Gemma 2 9B & $0.393$ & $0.946$ & $0.229$ & $0.354$ & $0.168$ \\
\cline{2-7}
 & Llama 3.1 70B & $0.445$ & $0.922$ & $0.245$ & $0.370$ & $0.196$ \\
\cline{2-7}
 & Llama 3.1 8B & $0.375$ & $0.948$ & $0.227$ & $0.349$ & $0.157$ \\
\cline{2-7}
 & Mistral 7B & $0.219$ & $0.993$ & $0.191$ & $0.306$ & $0.037$ \\
\cline{2-7}
 & Average & $0.358$ & $0.952$ & $0.223$ & $0.345$ & $0.140$ \\
\hline
\hline
\multirow{5}{*}{\textbf{F}} & Gemma 2 9B & $0.435$ & $0.922$ & $0.240$ & $0.364$ & $0.190$ \\
\cline{2-7}
 & Llama 3.1 70B & $0.444$ & $0.924$ & $0.246$ & $0.371$ & $0.200$ \\
\cline{2-7}
 & Llama 3.1 8B & $0.447$ & $0.909$ & $0.242$ & $0.366$ & $0.193$ \\
\cline{2-7}
 & Mistral 7B & $0.259$ & $0.986$ & $0.196$ & $0.313$ & $0.074$ \\
\cline{2-7}
 & Average & $\mathbf{0.396}$ & $0.935$ & $0.231$ & $\mathbf{0.354}$ & $\mathbf{0.164}$ \\
\hline
\end{tabular}
\end{table}

{\small\tabcolsep=3pt
\begin{longtable}{|c|c|c|c|c|c|c|c|c|}
\caption{Performance scores for the Output experiment macro-averaged over all datasets.}\label{tab:dcr.output.avg} \\
\endfirsthead
\hline
Prompt & Output & Model & Acc & Rec & Prec & F1 & MCC & AUPRC \\
\hline\hline
\endhead
\endfoot
\endlastfoot
\hline
Prompt & Output & Model & Acc & Rec & Prec & F1 & MCC & AUPRC \\
\hline\hline
\multirow{5}{*}{B} & \multirow{5}{*}{Word} & Gemma 9B & $0.440$ & $0.916$ & $0.242$ & $0.367$ & $0.193$ & $0.236$ \\
\cline{3-9}
 &  & Llama 70B & $0.520$ & $0.869$ & $0.266$ & $0.389$ & $0.230$ & $0.258$ \\
\cline{3-9}
 &  & Llama 8B & $0.397$ & $0.951$ & $0.228$ & $0.353$ & $0.172$ & $0.226$ \\
\cline{3-9}
 &  & Mistral 7B & $0.225$ & $0.993$ & $0.191$ & $0.307$ & $0.057$ & $0.191$ \\
\cline{3-9}
 &  & Average & $0.396$ & $0.932$ & $0.232$ & $0.354$ & $0.163$ & $0.228$ \\
\hline
\hline
\multirow{5}{*}{B} & \multirow{5}{*}{JSON} & Gemma 9B & $0.439$ & $0.900$ & $0.235$ & $0.358$ & $0.181$ & $0.230$ \\
\cline{3-9}
 &  & Llama 70B & $0.439$ & $0.929$ & $0.244$ & $0.369$ & $0.196$ & $0.241$ \\
\cline{3-9}
 &  & Llama 8B & $0.518$ & $0.802$ & $0.246$ & $0.360$ & $0.181$ & $0.237$ \\
\cline{3-9}
 &  & Mistral 7B & $0.227$ & $0.976$ & $0.192$ & $0.306$ & $0.049$ & $0.192$ \\
\cline{3-9}
 &  & Average & $0.406$ & $0.902$ & $0.229$ & $0.348$ & $0.152$ & $0.225$ \\
\hline
\hline
\multirow{5}{*}{B} & \multirow{5}{*}{Score} & Gemma 9B & $0.576$ & $0.652$ & $0.253$ & $0.342$ & $0.148$ & $0.266$ \\
\cline{3-9}
 &  & Llama 70B & $0.419$ & $0.942$ & $0.240$ & $0.366$ & $0.187$ & $0.286$ \\
\cline{3-9}
 &  & Llama 8B & $0.382$ & $0.946$ & $0.221$ & $0.343$ & $0.156$ & $0.240$ \\
\cline{3-9}
 &  & Mistral 7B & $0.196$ & $\mathbf{0.999}$ & $0.188$ & $0.302$ & $0.018$ & $0.204$ \\
\cline{3-9}
 &  & Average & $0.393$ & $0.885$ & $0.225$ & $0.338$ & $0.127$ & $0.249$ \\
\hline
\hline
\multirow{5}{*}{\textbf{B}} & \multirow{5}{*}{\textbf{Token-Word}} & Gemma 9B & $0.487$ & $0.876$ & $0.255$ & $0.378$ & $0.209$ & $0.403$ \\
\cline{3-9}
 &  & Llama 70B & $0.479$ & $0.904$ & $0.255$ & $0.380$ & $0.217$ & $0.376$ \\
\cline{3-9}
 &  & Llama 8B & $0.449$ & $0.918$ & $0.243$ & $0.366$ & $0.194$ & $0.366$ \\
\cline{3-9}
 &  & Mistral 7B & $0.236$ & $0.991$ & $0.193$ & $0.308$ & $0.066$ & $0.308$ \\
\cline{3-9}
 &  & Average & $0.413$ & $0.922$ & $0.236$ & $0.358$ & $0.171$ & $\mathbf{0.363}$ \\
\hline
\hline
\multirow{5}{*}{B} & \multirow{5}{*}{Token-JSON} & Gemma 9B & $0.447$ & $0.902$ & $0.240$ & $0.364$ & $0.190$ & $0.350$ \\
\cline{3-9}
 &  & Llama 70B & $0.405$ & $0.945$ & $0.237$ & $0.361$ & $0.178$ & $0.359$ \\
\cline{3-9}
 &  & Llama 8B & $0.554$ & $0.766$ & $0.261$ & $0.369$ & $0.194$ & $0.342$ \\
\cline{3-9}
 &  & Mistral 7B & $0.203$ & $0.998$ & $0.189$ & $0.303$ & $0.032$ & $0.327$ \\
\cline{3-9}
 &  & Average & $0.402$ & $0.903$ & $0.232$ & $0.349$ & $0.148$ & $0.344$ \\
\hline
\hline
\multirow{5}{*}{F} & \multirow{5}{*}{Word} & Gemma 9B & $0.435$ & $0.922$ & $0.240$ & $0.364$ & $0.190$ & $0.237$ \\
\cline{3-9}
 &  & Llama 70B & $0.444$ & $0.924$ & $0.246$ & $0.371$ & $0.200$ & $0.242$ \\
\cline{3-9}
 &  & Llama 8B & $0.447$ & $0.909$ & $0.242$ & $0.366$ & $0.193$ & $0.240$ \\
\cline{3-9}
 &  & Mistral 7B & $0.259$ & $0.986$ & $0.196$ & $0.313$ & $0.074$ & $0.196$ \\
\cline{3-9}
 &  & Average & $0.396$ & $0.935$ & $0.231$ & $0.354$ & $0.164$ & $0.229$ \\
\hline
\hline
\multirow{5}{*}{F} & \multirow{5}{*}{JSON} & Gemma 9B & $0.455$ & $0.905$ & $0.244$ & $0.368$ & $0.196$ & $0.241$ \\
\cline{3-9}
 &  & Llama 70B & $0.436$ & $0.932$ & $0.245$ & $0.371$ & $0.201$ & $0.241$ \\
\cline{3-9}
 &  & Llama 8B & $0.583$ & $0.780$ & $0.277$ & $0.388$ & $0.229$ & $0.262$ \\
\cline{3-9}
 &  & Mistral 7B & $0.510$ & $0.790$ & $0.248$ & $0.358$ & $0.170$ & $0.241$ \\
\cline{3-9}
 &  & Average & $0.496$ & $0.852$ & $0.254$ & $\mathbf{0.371}$ & $0.199$ & $0.246$ \\
\hline
\pagebreak
\multirow{5}{*}{F} & \multirow{5}{*}{Score} & Gemma 9B & $0.397$ & $0.927$ & $0.226$ & $0.348$ & $0.150$ & $0.295$ \\
\cline{3-9}
 &  & Llama 70B & $0.436$ & $0.932$ & $0.244$ & $0.370$ & $0.197$ & $0.264$ \\
\cline{3-9}
 &  & Llama 8B & $0.355$ & $0.953$ & $0.221$ & $0.343$ & $0.137$ & $0.285$ \\
\cline{3-9}
 &  & Mistral 7B & $0.196$ & $\mathbf{0.999}$ & $0.188$ & $0.303$ & $0.019$ & $0.216$ \\
\cline{3-9}
 &  & Average & $0.346$ & $\mathbf{0.953}$ & $0.220$ & $0.341$ & $0.126$ & $0.265$ \\
\hline
\hline
\multirow{5}{*}{F} & \multirow{5}{*}{Token-Word} & Gemma 9B & $0.447$ & $0.916$ & $0.242$ & $0.367$ & $0.194$ & $\mathbf{0.406}$ \\
\cline{3-9}
 &  & Llama 70B & $0.444$ & $0.918$ & $0.244$ & $0.369$ & $0.194$ & $0.365$ \\
\cline{3-9}
 &  & Llama 8B & $0.497$ & $0.883$ & $0.255$ & $0.378$ & $0.216$ & $0.360$ \\
\cline{3-9}
 &  & Mistral 7B & $0.294$ & $0.973$ & $0.202$ & $0.321$ & $0.094$ & $0.314$ \\
\cline{3-9}
 &  & Average & $0.420$ & $0.922$ & $0.236$ & $0.359$ & $0.175$ & $0.361$ \\
\hline
\hline
\multirow{5}{*}{F} & \multirow{5}{*}{Token-JSON} & Gemma 9B & $0.485$ & $0.888$ & $0.251$ & $0.375$ & $0.208$ & $0.368$ \\
\cline{3-9}
 &  & Llama 70B & $0.461$ & $0.895$ & $0.247$ & $0.370$ & $0.198$ & $0.362$ \\
\cline{3-9}
 &  & Llama 8B & $0.660$ & $0.701$ & $\mathbf{0.308}$ & $\mathbf{0.403}$ & $\mathbf{0.259}$ & $0.350$ \\
\cline{3-9}
 &  & Mistral 7B & $\mathbf{0.715}$ & $0.413$ & $0.289$ & $0.313$ & $0.157$ & $0.314$ \\
\cline{3-9}
 &  & Average & $\mathbf{0.580}$ & $0.724$ & $\mathbf{0.274}$ & $0.365$ & $\mathbf{0.206}$ & $0.349$ \\
\hline
\end{longtable}
}

\begin{table}[!htb]
\caption{Performance scores for the Few-Shot-Count experiment macro-averaged over all datasets.}\label{tab:dcr.n_shot.avg}
\resizebox{\linewidth}{!}{%
\centering
\begin{tabular}{|c|c|c|c|c|c|c|c|}
\hline
Examples & Model & Acc & Rec & Prec & F1 & MCC & AUPRC \\
\hline\hline
\multirow{5}{*}{0} & Gemma 2 9B & $0.487$ & $0.876$ & $0.255$ & $0.378$ & $0.209$ & $0.403$ \\
\cline{2-8}
 & Llama 3.1 70B & $0.479$ & $0.904$ & $0.255$ & $0.380$ & $0.217$ & $0.376$ \\
\cline{2-8}
 & Llama 3.1 8B & $0.449$ & $0.918$ & $0.243$ & $0.366$ & $0.194$ & $0.366$ \\
\cline{2-8}
 & Mistral 7B & $0.236$ & $0.991$ & $0.193$ & $0.308$ & $0.066$ & $0.308$ \\
\cline{2-8}
 & Average & $0.413$ & $0.922$ & $0.236$ & $0.358$ & $0.171$ & $0.363$ \\
\hline
\hline
\multirow{5}{*}{1} & Gemma 2 9B & $0.545$ & $0.840$ & $0.271$ & $0.392$ & $0.232$ & $0.438$ \\
\cline{2-8}
 & Llama 3.1 70B & $0.457$ & $0.937$ & $0.251$ & $0.378$ & $0.216$ & $0.443$ \\
\cline{2-8}
 & Llama 3.1 8B & $\mathbf{0.653}$ & $0.758$ & $\mathbf{0.306}$ & $\mathbf{0.421}$ & $\mathbf{0.277}$ & $0.470$ \\
\cline{2-8}
 & Mistral 7B & $0.223$ & $0.989$ & $0.191$ & $0.306$ & $0.048$ & $0.360$ \\
\cline{2-8}
 & Average & $\mathbf{0.470}$ & $0.881$ & $\mathbf{0.255}$ & $\mathbf{0.374}$ & $\mathbf{0.193}$ & $0.428$ \\
\hline
\hline
\multirow{5}{*}{2} & Gemma 2 9B & $0.522$ & $0.862$ & $0.264$ & $0.387$ & $0.225$ & $0.444$ \\
\cline{2-8}
 & Llama 3.1 70B & $0.451$ & $0.938$ & $0.249$ & $0.376$ & $0.212$ & $0.468$ \\
\cline{2-8}
 & Llama 3.1 8B & $0.492$ & $0.895$ & $0.252$ & $0.378$ & $0.211$ & $0.482$ \\
\cline{2-8}
 & Mistral 7B & $0.205$ & $0.994$ & $0.189$ & $0.304$ & $0.035$ & $0.414$ \\
\cline{2-8}
 & Average & $0.418$ & $0.922$ & $0.239$ & $0.361$ & $0.171$ & $0.452$ \\
\hline
\hline
\multirow{5}{*}{\textbf{3}} & Gemma 2 9B & $0.515$ & $0.871$ & $0.261$ & $0.384$ & $0.222$ & $0.455$ \\
\cline{2-8}
 & Llama 3.1 70B & $0.455$ & $0.927$ & $0.249$ & $0.374$ & $0.207$ & $0.481$ \\
\cline{2-8}
 & Llama 3.1 8B & $0.432$ & $0.935$ & $0.240$ & $0.366$ & $0.189$ & $\mathbf{0.490}$ \\
\cline{2-8}
 & Mistral 7B & $0.198$ & $0.997$ & $0.189$ & $0.303$ & $0.028$ & $0.446$ \\
\cline{2-8}
 & Average & $0.400$ & $0.932$ & $0.235$ & $0.357$ & $0.162$ & $\mathbf{0.468}$ \\
\hline
\hline
\multirow{5}{*}{4} & Gemma 2 9B & $0.526$ & $0.873$ & $0.265$ & $0.388$ & $0.230$ & $0.459$ \\
\cline{2-8}
 & Llama 3.1 70B & $0.463$ & $0.920$ & $0.253$ & $0.377$ & $0.211$ & $0.486$ \\
\cline{2-8}
 & Llama 3.1 8B & $0.414$ & $0.940$ & $0.235$ & $0.361$ & $0.176$ & $0.474$ \\
\cline{2-8}
 & Mistral 7B & $0.196$ & $\mathbf{0.998}$ & $0.189$ & $0.303$ & $0.025$ & $0.446$ \\
\cline{2-8}
 & Average & $0.400$ & $\mathbf{0.933}$ & $0.235$ & $0.357$ & $0.160$ & $0.466$ \\
\hline
\end{tabular}
}
\end{table}

\begin{table}[!htb]
\caption{Performance scores for the Few-Shot-Selection experiment macro-averaged over all datasets.}\label{tab:dcr.select.avg}
\resizebox{\linewidth}{!}{%
\centering
\begin{tabular}{|c|c|c|c|c|c|c|c|}
\hline
Method & Model & Acc & Rec & Prec & F1 & MCC & AUPRC \\
\hline\hline
\multirow{5}{*}{random} & Gemma 2 9B & $\mathbf{0.532}$ & $0.858$ & $\mathbf{0.270}$ & $\mathbf{0.392}$ & $\mathbf{0.233}$ & $0.377$ \\
\cline{2-8}
 & Llama 3.1 70B & $0.439$ & $0.934$ & $0.246$ & $0.372$ & $0.202$ & $0.359$ \\
\cline{2-8}
 & Llama 3.1 8B & $0.462$ & $0.876$ & $0.242$ & $0.364$ & $0.184$ & $0.351$ \\
\cline{2-8}
 & Mistral 7B & $0.214$ & $0.996$ & $0.191$ & $0.307$ & $0.058$ & $0.340$ \\
\cline{2-8}
 & Average & $\mathbf{0.412}$ & $0.916$ & $\mathbf{0.238}$ & $0.359$ & $\mathbf{0.169}$ & $0.357$ \\
\hline
\hline
\multirow{5}{*}{minilm} & Gemma 2 9B & $0.515$ & $0.872$ & $0.261$ & $0.384$ & $0.223$ & $0.455$ \\
\cline{2-8}
 & Llama 3.1 70B & $0.456$ & $0.927$ & $0.249$ & $0.374$ & $0.208$ & $0.482$ \\
\cline{2-8}
 & Llama 3.1 8B & $0.432$ & $0.935$ & $0.240$ & $0.366$ & $0.189$ & $0.490$ \\
\cline{2-8}
 & Mistral 7B & $0.198$ & $0.997$ & $0.189$ & $0.303$ & $0.028$ & $0.446$ \\
\cline{2-8}
 & Average & $0.400$ & $0.932$ & $0.235$ & $0.357$ & $0.162$ & $0.468$ \\
\hline
\hline
\multirow{5}{*}{\textbf{distroberta}} & Gemma 2 9B & $0.523$ & $0.864$ & $0.264$ & $0.387$ & $0.226$ & $0.469$ \\
\cline{2-8}
 & Llama 3.1 70B & $0.455$ & $0.928$ & $0.250$ & $0.375$ & $0.208$ & $0.485$ \\
\cline{2-8}
 & Llama 3.1 8B & $0.437$ & $0.927$ & $0.239$ & $0.365$ & $0.188$ & $\mathbf{0.495}$ \\
\cline{2-8}
 & Mistral 7B & $0.197$ & $\mathbf{0.998}$ & $0.188$ & $0.303$ & $0.028$ & $0.456$ \\
\cline{2-8}
 & Average & $0.403$ & $0.929$ & $0.235$ & $0.358$ & $0.163$ & $\mathbf{0.476}$ \\
\hline
\hline
\multirow{5}{*}{mpnet} & Gemma 2 9B & $0.516$ & $0.873$ & $0.263$ & $0.386$ & $0.226$ & $0.461$ \\
\cline{2-8}
 & Llama 3.1 70B & $0.460$ & $0.936$ & $0.253$ & $0.379$ & $0.220$ & $0.486$ \\
\cline{2-8}
 & Llama 3.1 8B & $0.439$ & $0.940$ & $0.244$ & $0.371$ & $0.199$ & $0.485$ \\
\cline{2-8}
 & Mistral 7B & $0.198$ & $\mathbf{0.998}$ & $0.189$ & $0.303$ & $0.031$ & $0.448$ \\
\cline{2-8}
 & Average & $0.404$ & $0.937$ & $0.237$ & $\mathbf{0.360}$ & $\mathbf{0.169}$ & $0.470$ \\
\hline
\hline
\multirow{5}{*}{mpnet-bal} & Gemma 2 9B & $0.472$ & $0.898$ & $0.246$ & $0.369$ & $0.200$ & $0.393$ \\
\cline{2-8}
 & Llama 3.1 70B & $0.475$ & $0.931$ & $0.257$ & $0.384$ & $0.228$ & $0.408$ \\
\cline{2-8}
 & Llama 3.1 8B & $0.317$ & $0.961$ & $0.204$ & $0.323$ & $0.110$ & $0.395$ \\
\cline{2-8}
 & Mistral 7B & $0.199$ & $0.996$ & $0.189$ & $0.303$ & $0.027$ & $0.339$ \\
\cline{2-8}
 & Average & $0.366$ & $\mathbf{0.947}$ & $0.224$ & $0.345$ & $0.141$ & $0.384$ \\
\hline
\end{tabular}
}
\end{table}

\begin{table}[!htb]
\caption{Performance scores for the Advanced-Techniques experiment macro-averaged over all datasets.}\label{tab:dcr.advanced.avg}
\resizebox{\linewidth}{!}{%
\centering
\begin{tabular}{|c|c|c|c|c|c|c|c|}
\hline
Variant & Model & Acc & Rec & Prec & F1 & MCC & AUPRC \\
\hline\hline
\multirow{5}{*}{base-zs} & Gemma 2 9B & $0.487$ & $0.876$ & $0.255$ & $0.378$ & $0.209$ & $0.403$ \\
\cline{2-8}
 & Llama 3.1 70B & $0.479$ & $0.904$ & $0.255$ & $0.380$ & $0.217$ & $0.376$ \\
\cline{2-8}
 & Llama 3.1 8B & $0.449$ & $0.918$ & $0.243$ & $0.366$ & $0.194$ & $0.366$ \\
\cline{2-8}
 & Mistral 7B & $0.236$ & $0.991$ & $0.193$ & $0.308$ & $0.066$ & $0.308$ \\
\cline{2-8}
 & Average & $0.413$ & $0.922$ & $0.236$ & $0.358$ & $0.171$ & $0.363$ \\
\hline
\hline
\multirow{5}{*}{\textbf{base-fs}} & Gemma 2 9B & $0.523$ & $0.864$ & $0.264$ & $0.387$ & $0.226$ & $0.469$ \\
\cline{2-8}
 & Llama 3.1 70B & $0.455$ & $0.928$ & $0.250$ & $0.375$ & $0.208$ & $0.485$ \\
\cline{2-8}
 & Llama 3.1 8B & $0.437$ & $0.927$ & $0.239$ & $0.365$ & $0.188$ & $\mathbf{0.495}$ \\
\cline{2-8}
 & Mistral 7B & $0.197$ & $\mathbf{0.998}$ & $0.188$ & $0.303$ & $0.028$ & $0.456$ \\
\cline{2-8}
 & Average & $0.403$ & $\mathbf{0.929}$ & $0.235$ & $0.358$ & $0.163$ & $\mathbf{0.476}$ \\
\hline
\hline
\multirow{5}{*}{chu} & Gemma 2 9B & $0.500$ & $0.868$ & $0.255$ & $0.376$ & $0.208$ & $0.258$ \\
\cline{2-8}
 & Llama 3.1 70B & $0.479$ & $0.892$ & $0.251$ & $0.375$ & $0.209$ & $0.253$ \\
\cline{2-8}
 & Llama 3.1 8B & $0.569$ & $0.793$ & $0.271$ & $0.382$ & $0.222$ & $0.273$ \\
\cline{2-8}
 & Mistral 7B & $0.266$ & $0.981$ & $0.197$ & $0.314$ & $0.079$ & $0.202$ \\
\cline{2-8}
 & Average & $0.454$ & $0.884$ & $0.243$ & $0.362$ & $0.179$ & $0.246$ \\
\hline
\hline
\multirow{5}{*}{chu-fs} & Gemma 2 9B & $0.529$ & $0.846$ & $0.265$ & $0.384$ & $0.220$ & $0.270$ \\
\cline{2-8}
 & Llama 3.1 70B & $0.489$ & $0.908$ & $0.257$ & $0.383$ & $0.224$ & $0.263$ \\
\cline{2-8}
 & Llama 3.1 8B & $0.349$ & $0.930$ & $0.211$ & $0.328$ & $0.119$ & $0.217$ \\
\cline{2-8}
 & Mistral 7B & $0.315$ & $0.963$ & $0.208$ & $0.327$ & $0.116$ & $0.217$ \\
\cline{2-8}
 & Average & $0.421$ & $0.912$ & $0.235$ & $0.355$ & $0.170$ & $0.241$ \\
\hline
\hline
\multirow{5}{*}{cot1} & Gemma 2 9B & $0.542$ & $0.808$ & $0.263$ & $0.379$ & $0.210$ & $0.251$ \\
\cline{2-8}
 & Llama 3.1 70B & $0.497$ & $0.816$ & $0.249$ & $0.362$ & $0.171$ & $0.245$ \\
\cline{2-8}
 & Llama 3.1 8B & $0.610$ & $0.733$ & $0.282$ & $0.387$ & $0.225$ & $0.264$ \\
\cline{2-8}
 & Mistral 7B & $\mathbf{0.743}$ & $0.328$ & $\mathbf{0.306}$ & $0.290$ & $0.148$ & $0.247$ \\
\cline{2-8}
 & Average & $\mathbf{0.598}$ & $0.671$ & $\mathbf{0.275}$ & $0.354$ & $\mathbf{0.188}$ & $0.252$ \\
\hline
\hline
\multirow{5}{*}{cot2} & Gemma 2 9B & $0.487$ & $0.850$ & $0.247$ & $0.365$ & $0.190$ & $0.240$ \\
\cline{2-8}
 & Llama 3.1 70B & $0.480$ & $0.897$ & $0.254$ & $0.378$ & $0.214$ & $0.249$ \\
\cline{2-8}
 & Llama 3.1 8B & $0.602$ & $0.740$ & $0.291$ & $\mathbf{0.394}$ & $\mathbf{0.234}$ & $0.267$ \\
\cline{2-8}
 & Mistral 7B & $0.375$ & $0.851$ & $0.207$ & $0.318$ & $0.088$ & $0.205$ \\
\cline{2-8}
 & Average & $0.486$ & $0.834$ & $0.250$ & $\mathbf{0.364}$ & $0.181$ & $0.240$ \\
\hline
\hline
\multirow{5}{*}{tot} & Gemma 2 9B & $0.606$ & $0.657$ & $0.270$ & $0.364$ & $0.186$ & $0.245$ \\
\cline{2-8}
 & Llama 3.1 70B & $0.475$ & $0.884$ & $0.250$ & $0.372$ & $0.202$ & $0.244$ \\
\cline{2-8}
 & Llama 3.1 8B & $0.526$ & $0.703$ & $0.229$ & $0.328$ & $0.125$ & $0.223$ \\
\cline{2-8}
 & Mistral 7B & $0.712$ & $0.333$ & $0.265$ & $0.273$ & $0.109$ & $0.227$ \\
\cline{2-8}
 & Average & $0.580$ & $0.644$ & $0.253$ & $0.335$ & $0.155$ & $0.235$ \\
\hline
\end{tabular}
}
\end{table}

\clearpage

\begin{table}[!htb]
\caption{Performance scores for the Baseline experiment macro-averaged over all datasets.}\label{tab:dcr.final.avg}
\centering
\begin{tabular}{|c|c|c|c|c|c|c|}
\hline
Algorithm & Accuracy & Recall & Precision & F1 Score & MCC & AUPRC \\
\hline\hline
gemm & $0.523$ & $0.864$ & $0.264$ & $0.387$ & $0.226$ & $0.469$ \\
\hline
ll-70b & $0.455$ & $0.928$ & $0.250$ & $0.375$ & $0.208$ & $0.485$ \\
\hline
ll-8b & $0.437$ & $0.927$ & $0.239$ & $0.365$ & $0.188$ & $0.495$ \\
\hline
mistr & $0.197$ & $\mathbf{0.998}$ & $0.188$ & $0.303$ & $0.028$ & $0.456$ \\
\hline
gptmi & $0.546$ & $0.811$ & $0.272$ & $0.385$ & $0.221$ & $0.335$ \\
\hline
gptna & $0.711$ & $0.664$ & $0.338$ & $0.428$ & $0.291$ & $0.439$ \\
\hline
nb & $0.810$ & $0.648$ & $0.449$ & $\mathbf{0.517}$ & $\mathbf{0.413}$ & $0.499$ \\
\hline
rf-d & $0.746$ & $0.715$ & $0.381$ & $0.467$ & $0.361$ & $0.513$ \\
\hline
\textbf{rf-t} & $0.666$ & $0.838$ & $0.352$ & $0.458$ & $0.353$ & $\mathbf{0.556}$ \\
\hline
debe & $0.829$ & $0.413$ & $0.453$ & $0.419$ & $0.323$ & $0.474$ \\
\hline
scid & $\mathbf{0.834}$ & $0.505$ & $\mathbf{0.498}$ & $0.489$ & $0.396$ & $0.531$ \\
\hline
\end{tabular}
\end{table}

\section{Precision-recall curves}\label{app:prc}

The following figures contain the PR curves for Output, Few-Shot-Count, Few-Shot-Selection and Baseline experiments, which were omitted from the main manuscript.

\clearpage
\vspace*{-4cm}
\noindent
\hspace*{-2.85cm}
\includegraphics[scale=0.55]{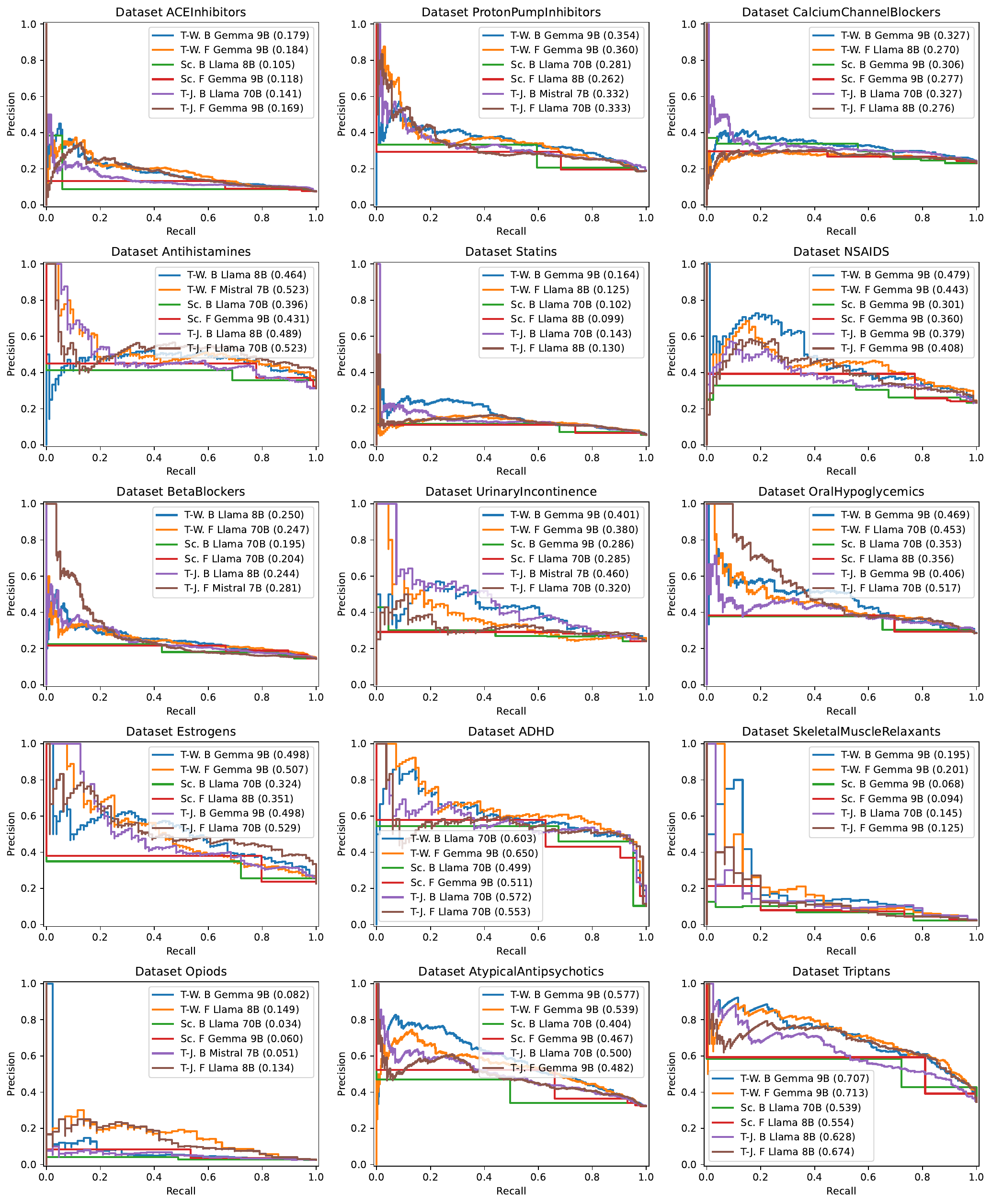}
\captionof{figure}{PR curves for the Output experiment.}
\label{fig:output:pr}

\clearpage
\vspace*{-4cm}
\noindent
\hspace*{-2.85cm}
\includegraphics[scale=0.55]{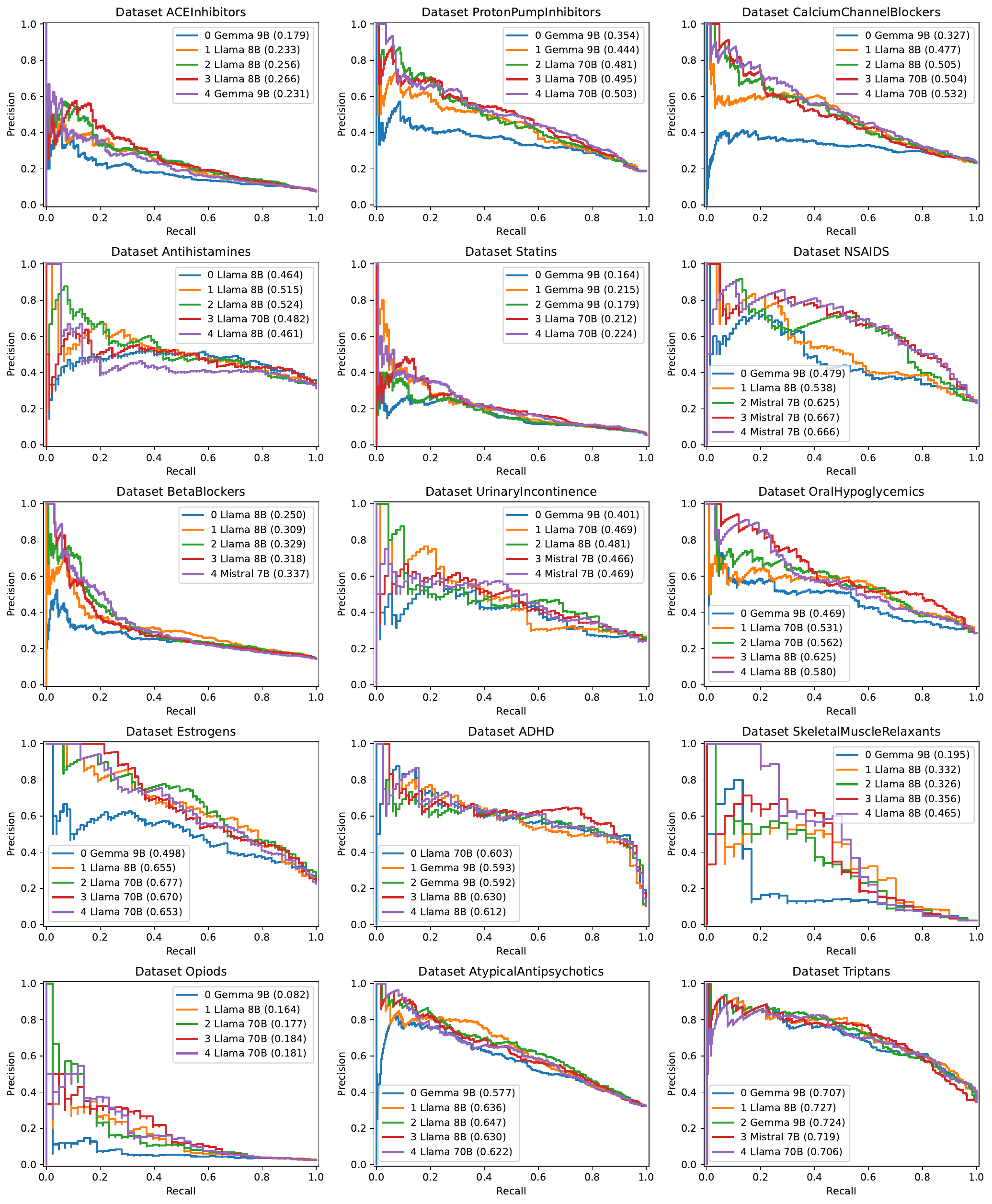}
\captionof{figure}{PR curves for the Few-Shot-Count experiment.}
\label{fig:n_shot:pr}

\clearpage
\vspace*{-4cm}
\noindent
\hspace*{-2.85cm}
\includegraphics[scale=0.55]{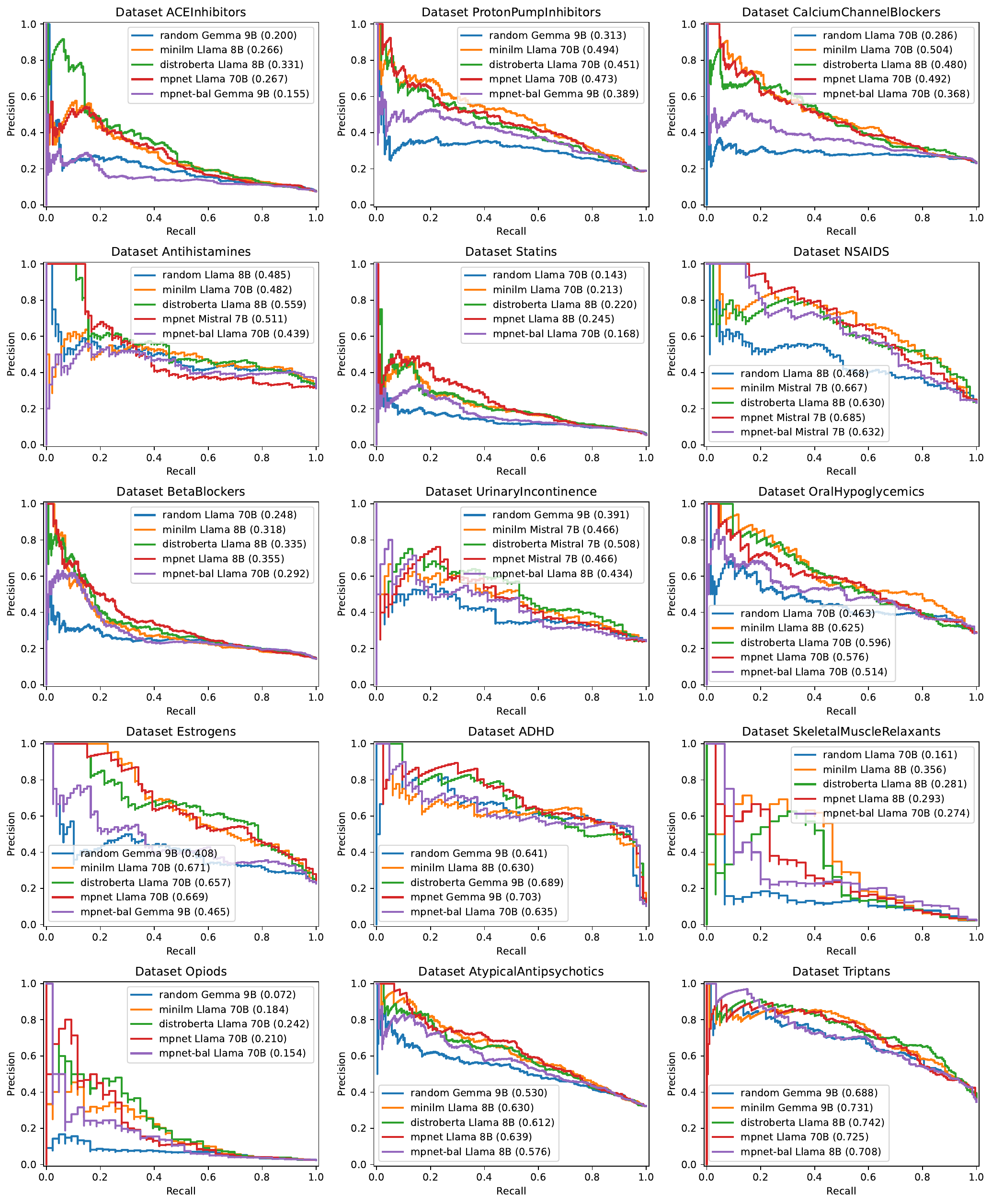}
\captionof{figure}{PR curves for the Few-Shot-Selection experiment.}
\label{fig:select:pr}

\clearpage
\vspace*{-4cm}
\noindent
\hspace*{-2.85cm}
\includegraphics[scale=0.55]{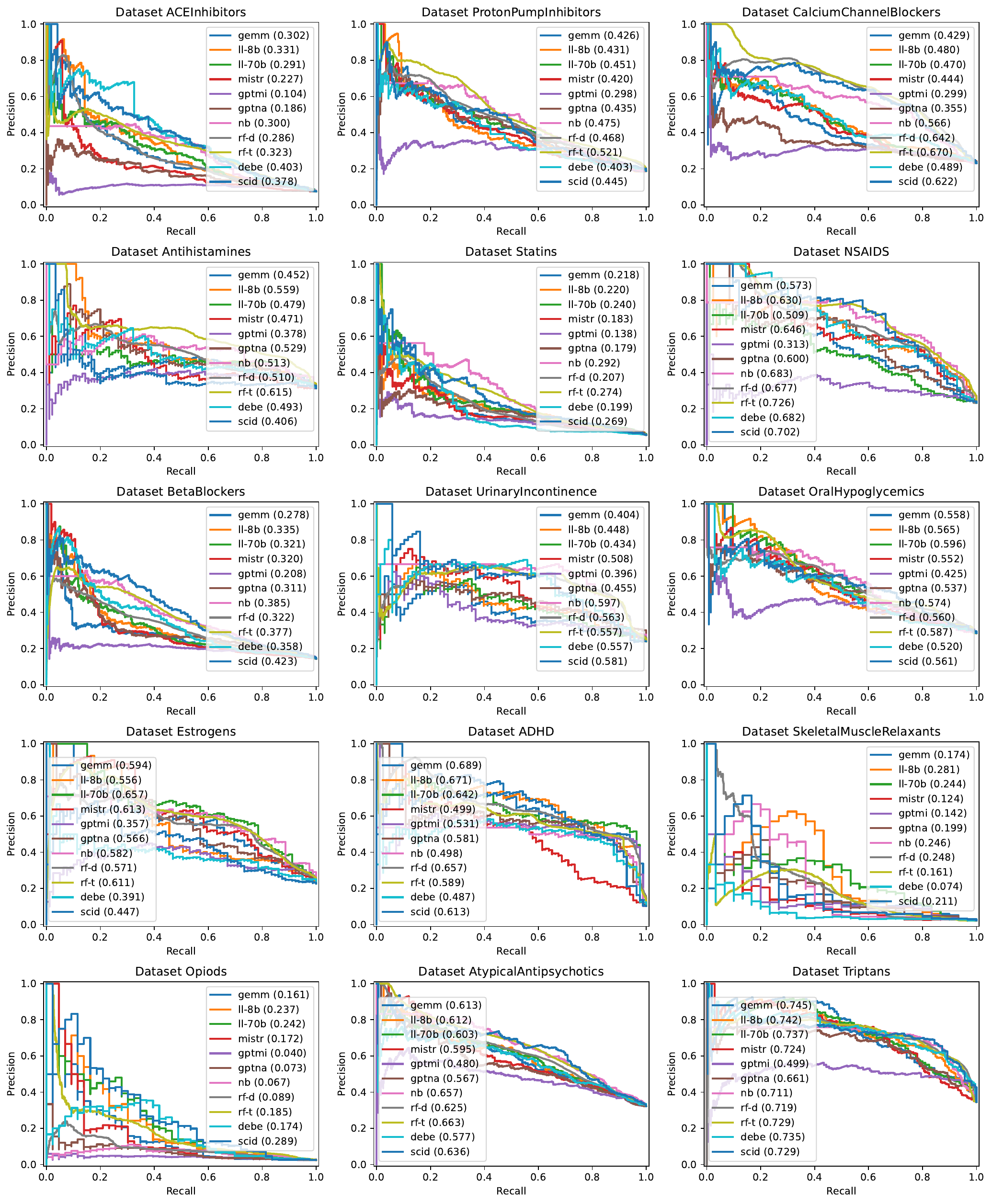}
\captionof{figure}{PR curves for the Baseline experiment.}
\label{fig:final:pr}

\section{Code, data and results availability}

The code used to run this article's experiments, the processed Systematic Drug Class Review datasets well as the results on various levels of aggregation are available on Zenodo: \url{https://zenodo.org/records/18874261}.

\end{document}